\documentclass[11pt,a4paper]{article}
\usepackage[table]{xcolor} 
\usepackage[hyperref,acceptedWithA]{tacl2018v2}
\usepackage{times}
\usepackage{latexsym}
\usepackage{multirow,bigdelim,dcolumn,booktabs,array}
\usepackage{graphicx}
\usepackage[shortlabels]{enumitem}
\usepackage{float}

\usepackage{todonotes}
\usepackage{booktabs}
\usepackage{dirtytalk}
\usepackage{microtype}
\usepackage{multirow}

\usepackage{amsmath}
\usepackage{amssymb}
\usepackage{subcaption}
\usepackage{xspace}
 
\usepackage{url}

\definecolor{very_likely}{HTML}{273253}
\definecolor{somewhat_likely}{HTML}{273253}
\definecolor{very_unlikely}{HTML}{c28004}
\definecolor{somewhat_unlikely}{HTML}{c28004}
\definecolor{background1}{HTML}{f7f7f7}

\definecolor{red}{HTML}{de2d26}
\definecolor{green}{HTML}{31a354}

\title{Joint Universal Syntactic and Semantic Parsing} 

\author{Elias Stengel-Eskin \\
  Johns Hopkins University  \\
\And
Kenton Murray \\
Johns Hopkins University  \\
\And 
  Sheng Zhang \\
  Microsoft Research \\
\AND
  Aaron Steven White  \\
  University of Rochester \\
\And
  Benjamin Van Durme \\
  Johns Hopkins University  \\ }

\date{}

\begin{document}
\maketitle
\begin{abstract} 
While numerous attempts have been made to jointly parse syntax and semantics, high performance in one domain typically comes at the price of performance in the other.
This trade-off contradicts the large body of research focusing on the rich interactions at the syntax-semantics interface. 
We explore multiple model architectures which allow us to exploit the rich syntactic and semantic annotations contained in the Universal Decompositional Semantics (UDS) dataset,
jointly parsing Universal Dependencies and UDS to obtain state-of-the-art results in both formalisms.
We analyze the behaviour of a joint model of syntax and semantics, finding patterns supported by linguistic theory at the syntax-semantics interface. 
We then investigate to what degree joint modeling generalizes to a multilingual setting, where we find similar trends across 8 languages. 
\end{abstract}

\section{Introduction} 
Given their natural expression in terms of hierarchical structures and their well-studied interactions,  syntax and semantics have long been treated as parsing tasks, both independently and jointly. 
One would expect joint models to outperform separate or pipelined ones; however, many previous attempts have yielded mixed results, finding that while one level can be used as an additional signal to benefit the other \citep{swayamdipta.s.2017, swayamdipta.s.2018, johansson.r.2008}, obtaining high performance in both syntax and semantics simultaneously is difficult \citep{krishnamurthy.j.2014,hajivc.j.2009}.  

A variety of tree- and graph-based representations have been devised for representing syntactic structure (e.g. varieties of constituency and dependency parse trees) as well as semantic structure, e.g. Abstract Meaning Representation \citep[AMR;][]{banarescu.l.2013}, Universal Conceptual Cognitive Annotation \citep[UCCA;][]{abend.o.2013}, and Semantic Dependency Parsing formalisms \citep[SDP;][]{oepen.s.2014, oepen.s.2015, oepen.s.2016}. 
These semantic representations have varying degrees of abstraction from the input and syntax, ranging from being directly tied to the input tokens (e.g. SDP formalisms) to being heavily abstracted away from it (e.g. AMR, UCCA). 
Universal Decompositional Semantics \citep[UDS;][]{white.a.2020} falls between these extremes, with a semantic graph that is closely tied to the syntax while not being constrained to match the input tokens. 
Crucially, UDS graphs not only represent the predicate-argument relationships in the input, but also host scalar-valued crowdsourced annotations encoding a variety of semantic inferences, described in \S\ref{sec:data}.
These provide another level for linguistic analysis and make UDS unique among meaning representations. 
Furthermore, as UDS graphs build on Universal Dependency (UD) parses, UDS is naturally positioned to take full advantage of the extensive and linguistically diverse set of UD annotations.

We extend the transductive sequence-to-graph UDS parser proposed by \citet{stengel-eskin.e.2020} to simultaneously perform Universal Dependencies (UD) and UDS parsing, finding that joint modeling offers concomitant benefits to both tasks. 
In particular, after exploring several multitask learning objectives and model architectures for integrating syntactic and semantic information, we obtain our best overall results with an encoder-decoder semantic parsing model, where the encoder is shared with a biaffine syntactic parser \citep{dozat.t.2016}.
Because the UDS dataset is annotated on an existing UD corpus, our experiments isolate the effect of adding a semantic signal without the confound of additional data: all our monolingual systems are trained on the same set of sentences. 

In contrast to previous work on joint syntax-semantics parsing, we are able to achieve high performance in both domains with a single, unified model; this is particularly salient to UDS parsing, as the UD parse is a central part of a \emph{complete} UDS analysis. 
Our best joint model's UDS performance beats the previous best by a large margin, while also yielding SOTA scores on a semantically valid subset of English Web Treebank (EWT). 
We introduce a model optimized for UD, which can obtain competitive UDS performance while matching the current SOTA UD parser on the whole of EWT.\footnote{Models/code: {\href{https://github.com/esteng/miso_uds}{github.com/esteng/miso\_uds}}}

We analyze these objectives and architectures with LSTM encoders and decoders as well as novel Transformer-based sequence-to-graph architectures, which outperform the LSTM variants. 
While previous work suggests that contextualized encoders largely obviate the need for an explicit syntactic signal in semantic tasks \citep{swayamdipta.s.2019, glavas.g.2020}, we find that syntactic (and semantic) annotations provide consistent performance gains even when such encoders are used. 
This suggests that the UD and UDS signals are complementary to the signal encoded by a pretrained encoder, and so we tune the encoder at various depths, further improving performance.


Building on this result, we leverage the shared multilingual representation space of XLM-R \citep{conneau.a.2019} to examine UD parsing in 8 languages across 5 families and varying typological settings, where we demonstrate a cross-lingual benefit of UDS parsing on UD parsing. 

\section{Background and Related Work} \label{sec:background}
In both language production and comprehension, syntax and semantics play complementary roles.
Their close relationship has also been noted in language acquisition research, with the ``semantic bootstrapping'' hypothesis proposing that infants use semantic role information as an inductive bias for acquiring syntax \citep{pinker.s.1979,pinker.s.1984},\footnote{\citet{abend.o.2017} presents an implementation of the semantic bootsrapping hypothesis.}
while \citet{landau.b.1985, gleitman.l.1990} and \citet{naigles.l.1990} present evidence that infants use syntactic information when acquiring novel word meanings. 
Their connection was codified in \citet{montague.r.1970}'s seminal work formalizing the link between syntactic structures and formal semantics. 
Broadly, their interactions can be split into ``bottom-up'' constraints on the semantics of an utterance from its syntax, and ``top-down'' constraints on the syntax, based on the semantics. 
Despite their close empirical and theoretical ties, work on  predicting syntax and semantic structures jointly has often struggled to attain high performance in one domain without compromising on performance in the other. 

\vspace{-0.5em} 
\paragraph{CCG-based parsing} Following in the Montagovian tradition, several computational formalisms and models have focused on the syntax-semantics interface, including the Head-Driven Phrase Structure Grammar \citep{pollard.c.1994} and Combinatory Categorical Grammar (CCG) \citep{steedman.m.2000}. In particular, CCG syntactic types can be paired with functional semantic types (e.g. $\lambda$ calculus strings) to compositionally construct logical forms. \citet{krishnamurthy.j.2014} model this process with a linear model over both syntactic derivations and logical forms, 
trained with a discriminative objective that combines direct syntactic supervision with distant supervision from the logical form, 
finding that while joint modeling is feasible, it slightly lowers the syntactic performance. 
By way of contrast, \citet{lewis.m.2015} find that a joint CCG and Semantic Role Labelling (SRL) dependency parser outperforms a pipeline baseline. 
The semantic signal can also be used to induce syntax without using syntactic supervision. 

\vspace{-0.5em}
\paragraph{AMR parsing} CCG approaches have also been applied to semantics-only AMR parsing   \citep{artzi.y.2015, misra.d.2016,  beschke.s.2019}.
Jointly modeling AMR and syntax, \citet{zhou.q.2020} induce a soft syntactic structure with a latent variable model, obtaining slight improvements over semantics-only models in low-resource settings. 

\vspace{-0.5em} 
\paragraph{SRL parsing} SRL dependency parsing---the task of labeling an utterance's predicates and their respective arguments in a possibly non-projective directed acyclic graph (DAG)---is more akin to the UDS parsing task than CCG parsing and has an equally robust foundation of empirical results, having been the focus of several CoNLL shared tasks---most relevantly the 2008 and 2009 shared tasks, which were on joint syntactic and SRL dependency parsing \citep{surdeanu.m.2008,hajivc.j.2009}. The upshot of these tasks was that a joint syntactic and semantic analysis could provide benefits over a separated system \citep{johansson.r.2008} but that in a multilingual setting, SRL-only systems slightly outperformed joint systems on average \citep{hajivc.j.2009}. While the systems presented in these challenges used hand-crafted features, \citet{swayamdipta.s.2016}  replicated many of their results in a neural setting. 
For SRL tagging, \citet{strubell.e.2018} introduce an end-to-end neural model that also uses UD parsing as a multitask intermediate task, akin to our intermediate model. 

\vspace{-0.5em}
\paragraph{Syntactic scaffolds} Like our models, work on syntactic scaffolds introduces a multitask learning \citep{caruana.r.1997} framework, where a syntactic auxiliary task is introduced for the benefit of a semantic task;
in contrast to the systems presented here, the syntactic task is treated as a purely auxiliary signal, with the model evaluation coming solely from the semantic task.
\citet{swayamdipta.s.2017} first introduce the notion of a syntactic scaffold for frame-semantic parsing, where a lightweight syntactic task (constituent labeling) is used as an auxiliary signal in a multitask learning setup to the benefit of the semantic task. \citet{swayamdipta.s.2018} introduce a similar syntactic scaffolding objective for three semantic tasks.
However, \citet{swayamdipta.s.2019} find that the benefits of shallow syntactic objectives are largely eclipsed by the implicit information captured in contextualized encoders. 

\section{Data}\label{sec:data} 
\begin{figure}[h]
    \centering
    \includegraphics[width=\linewidth]{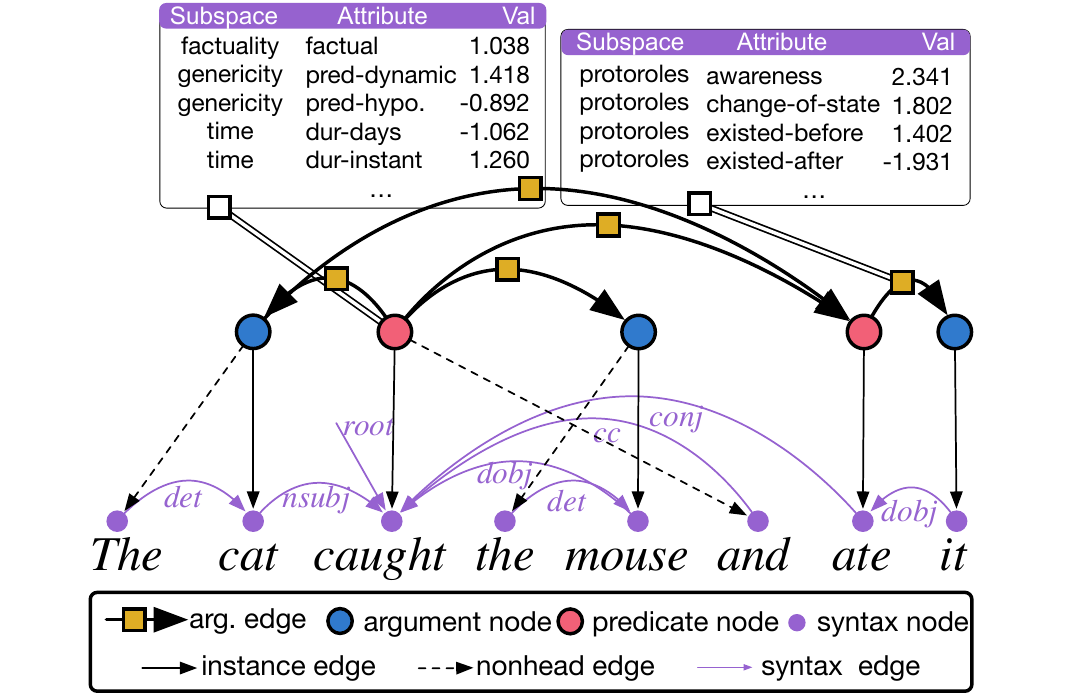}
    \caption{A UDS graph with syntactic and semantic parses, as well as node- and edge-level properties.}
    \label{fig:decomp_graph}
    \vspace{-1.25em}
\end{figure}

A number of factors make the Universal Decompositional Semantics representation \citep[UDS;][]{white.a.2020} particularly well-suited to our purposes, especially the existence of parallel manually-annotated syntactic and semantic data. 
In UDS, a semantic graph is built on top of existing English Web Treebank \citep[EWT;][]{bies.a.2012} UD parses, which are mapped to nodes and edges in a semantic DAG via a set of deterministic rules  \citep{white.a.2016,zhang.s.2017}. 
This semantic graph, which represents the predicate-argument relationships in the text, is then annotated with crowdsourced scalar-valued attributes falling into the following categories: 
factuality (how likely a predicate is to have occurred), genericity (how general or specific a predicate/argument is), time (how long an event took), word sense (which word senses apply to a predicate/argument) and semantic proto-roles, which break the traditional SRL ontologies into simpler ``proto-agent'' properties (e.g. volition, awareness, sentience) and ``proto-patient'' properties (e.g. change of state, change of location, being used). 
Note that while the semantic graph structure is tied to the syntax, the attribute values, encoding fine-grained, abstracted semantic inferences, are not. These attributes are unique among graph-based semantic representations. 
All of these properties are annotated on a scale of $-3$ to $3$; for more details on the dataset, we refer the reader to \citet{white.a.2020} and \citet{stengel-eskin.e.2020}, as well as Fig.~\ref{fig:decomp_graph}. 
We train and evaluate on a semantically valid subset of EWT.\footnote{This excludes roughly 20\% of the EWT data; excluded sentences include forms of address (e.g. ``Dear Nina ,'') , URLs, and discourse markers (e.g. ``( Applause . )'') which lack predicate-argument structures.}
We similarly limit our baselines to these examples for our UD analysis, and release our cleaned UD dataset, for which we report state-of-the-art parsing performance.

\begin{figure}[h]
    \centering
    \vspace{-1em}
    \includegraphics[width=\columnwidth]{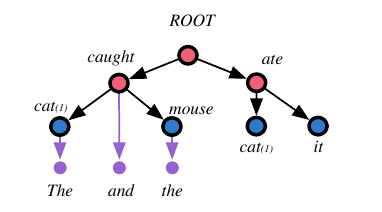}
    \vspace{-2em}
    \caption{Conversion of Fig.~\ref{fig:decomp_graph} to an arborescence.}
    \label{fig:arborescence}
    \vspace{-1.5em}
\end{figure}

\paragraph{Arborescence} Following \citet{zhang.s.2019a} and \citet{stengel-eskin.e.2020}, we convert the UDS graph into an arborescence, or tree.
Re-entrant semantic nodes are copied and co-indexed, so that a DAG can be recovered via a deterministic post-processing step. 
Each node is assigned a token label, taken from the token corresponding to the syntactic head of the semantic node (instance edges in Fig.~\ref{fig:decomp_graph}). 
All syntactic nodes not assigned as labels to semantic nodes are included as nodes dominated by their corresponding semantic node (pink edges in Fig.~\ref{fig:arborescence}).\footnote{``and'' is assigned to ``caught'' based on it being the head of the {\tt{cc}} relation in the UD parse}
An example conversion of the UDS graph in Fig.~\ref{fig:decomp_graph} can be found in Fig.~\ref{fig:arborescence}. 
In the ``semantics only'' setting, only the semantic nodes are included in the graph. 
A pre-order traversal linearizes the arborescence into a sequence of nodes, indices, edge heads and labels, and the corresponding node and edge attributes. 

\section{Models}\label{sec:model} We build on the transductive parsing model presented by \citet{stengel-eskin.e.2020}, which itself builds on the broad-coverage semantic parsing model of \citet{zhang.s.2019b} and relies heavily on AllenNLP \citep{gardner.m.2017}. 
The transductive parsing paradigm recasts graph-based parsing as a sequence-to-graph problem, using an attentional sequence-to-sequence model to transduce the input sentence into a set of nodes while incrementally predicting edge and edge labels for those nodes. 
The UDS semantic parser consists of 7 modules:  

\noindent\textbf{The encoder module} embeds the input features (type-level GloVe and contextualized word embeddings, POS tags, charCNN features) into a latent space, producing one vector per input token. BERT representations are pooled over subword units.\footnote{This input is augmented with a sinusoidal position embedding for the Transformer models.}

\noindent\textbf{The decoder embedding module} embeds the categorical information of the previous timestep (e.g. the token identity and index, the head token identity and index, the edge type) into a real space. 

\noindent\textbf{The target node module} builds node representations from the decoder embedding module's output. 

\noindent\textbf{The target label module} extends the Pointer-Generator network \citep{see.a.2017}, which supports both generating new token labels from a vocabulary and copying tokens from the input, with a ``target-copy'' operation, additionally allowing the model to predict a token label by copying a previously predicted node, conditioned on a target node. 
This three-operation approach (i.e. generate, source-copy, target-copy) enables the parser to seamlessly handle lexicalized and non-lexicalized formalisms, while also natively supporting re-entrancy through the target-copy operation.

\noindent\textbf{The relation module} is a graph-based dependency parser based on the parser presented by \citet{dozat.t.2016} which uses separate head and dependency MLPs followed by a biaffine transformation to predict the dependency scores and labels between each node in a fully connected graph.
For semantic parsing, we follow a greedy decoding strategy since the linearization of the arborescence implicitly enforces a well-formed output; this allows for single-step online decoding.

\noindent\textbf{The node attribute module} uses the node representations to predict  whether each attribute applies to each node, and what its value should be. 
Deciding whether the attribute applies and the prediction of its value are performed by separate MLPs. 

\noindent\textbf{The edge attribute module} is similar to the node attribute module, but passes a bilinear transformation of two node representations to the MLPs, which predict edge-level properties and masks. 

In this work, we make several modifications  to this model. 
Firstly, we replace the encoder and target node modules with Transformer-based architectures \citep{vaswani.a.2017}. 
Next, we introduce a second biaffine parser on top of the output of the encoder module, which is tasked with performing dependency parsing on the UD data. 
During training, we use greedy decoding, while at test time the Chu-Liu-Edmonds Maximum Spanning Tree algorithm \citep{chu.l.1965,edmonds.j.1967} is used.

More specifically, the encoded input representations $\mathbf{s}_t$ in \citet{stengel-eskin.e.2020} were obtained from a stacked bidirectional LSTM.  
While the input to this module remains the same in our Transformer-based implementation, the representation $\mathbf{s}_t$ is now given by the final layer of a multi-head multi-layer Transformer model. 
Crucially, following \citet{nguyen.t.2019}
we replace the layer normalization layer with a ScaleNorm layer and place it \emph{before} the feed-forward network.
The syntactic biaffine parser uses the encoder representations $\mathbf{s}$ to predict a head and head label for each token. This task is different from the semantic parsing task as the graph is lexicalized (i.e. there is a bijection between input tokens and graph nodes). 

Similarly, the node representations $\mathbf{z}_i$ are computed by a Transformer decoder with both self-attention (as in the encoder) and source-side attention. 
The first layer of node representations for the decoder is given by learned continuous embeddings of the head token and current token representation, their respective indices, and the relationship between them. 
During training, the gold nodes and heads are used (i.e. teacher forcing), and the attention 
is computed with an autoregressive mask, so that each token is only able to attend to tokens in its left context. 
We take $\mathbf{z}_i = \text{ScaleNorm}(\mathbf{x}^L_i)$, $\mathbf{x}_i^L$ being the output of the final layer in a stack of $L$ Transformer layers.
We also follow \citeauthor{nguyen.t.2019} in scaling the attention head weight initialization by a factor of $k$.

\section{Experiment 1: Joint English Parsing} To determine the effect of jointly parsing the syntax and semantics, we consider a number of baselines and experimental settings. First, we contrast our re-implementation of \citet{stengel-eskin.e.2020}'s LSTM-based model with their results.
We then report the results of our Transformer-based model, described in \S\ref{sec:model}. 
After establishing these baselines, we consider different methods of incorporating the syntactic signal into the model:

\noindent\textbf{Concat-before (CB):} Here, we linearize the syntactic UD parse, which is a tree, via a pre-order traversal, yielding a sequence of nodes, edge heads, and edge labels, which we prepend to the corresponding sequences obtained by linearizing the semantic parse, separated by a special separation token. At inference time, we use this token to split the output into syntactic and semantic parses.  

\noindent\textbf{Concat-after (CA):} This setting is identical to the concat-before setting, except that the syntactic graph is appended at the end of the semantic sequence. These two settings incorporate some syntactic signal into the semantic parse, but do not exploit UD parsing's lexicalization assumption; thus we expect them to yield subpar UD results. 

\noindent\textbf{Encoder-side (EN):} \label{sec:syn_model} We do make use of this assumption here, adding a biaffine parser to the encoder states $\mathbf{s}_{1:T}$. 
We introduce an additional syntactic parsing objective, which allows us to take advantage of the strong lexicalized bias. 
However, the syntactic signal only enters the model implicitly via backprop, i.e. during the forward pass, the model has no access to syntactic information. 

\noindent\textbf{Intermediate (IN):} We incorporate the syntactic information by re-encoding the predicted syntactic parse and passing it to the decoder. Due to the close syntactic correspondence of the UDS semantic graph, we would expect that allowing the decoder to access the predicted dependency parse would benefit both the semantic parse as well as the syntactic parse.
We enable this by concatenating edge information to $\mathbf{s}_{1:T}$ and linearly projecting it. 
Specifically, given edge head scores $\mathbf{E} \in \mathbb{R}^{T \times T}$, where each row $i$ is a distribution over possible heads for token $i$, the output of the parser's head MLP $\mathbf{H} \in \mathbb{R}^{T \times d_h}$, and the output of the parser's edge type MLP $\mathbf{T} \in \mathbb{R}^{T \times d_t}$, we compute the new encoder representations $\mathbf{s}'$ as: 
\vspace{-0.5em}
\begin{align*}
    \mathbf{H}' &= \mathbf{H}^T \mathbf{E}\:,\:\mathbf{T}' = \mathbf{T}^T \mathbf{E}\\
    \mathbf{s}'_i &= [\mathbf{s}_i; \mathbf{H}'_{i}; \mathbf{T}'_{i}]^T \mathbf{W}^I, \mathbf{W}^I \in \mathbb{R}^{(d_{s} + d_{h} + d_{t}) \times d_{s}}
\end{align*}
\paragraph{Transformer Hyperparameters} Unlike the LSTM-based model, which is fairly robust to hyperparameter changes, the Transformer-based architecture was found to be sensitive to such changes. We use a random search strategy \citep{bergstra.j.2012} with 40 replicants, tuning the number of layers $l \in [6, 8, 12]$, the initialization scaling factor $k \in [4, 32, 128, 512]$, the number of heads $H \in [4,8]$, the dropout factor $d \in [0.20, 0.33]$, and the number of warmup steps for the optimizer $w \in [1000, 4000, 8000]$. This was performed with the base model, with the best hyperparameters used in all other models. 

\subsection{Evaluation Metrics}

\noindent\textbf{UAS/LAS:} Unlabeled Attachment Score (UAS) computes the fraction of tokens with correctly assigned heads in a dependency parse. 
Labeled Attachment Score (LAS) computes the fraction with correct heads and arc labels. 
Both are standard metrics for UD parsing. 

\noindent\textbf{Pearson's $\rho$:} For UDS attributes, we compute the Pearson correlation between the predicted attributes at each node and the gold annotations in the UDS corpus. 
This is obtained under an ``oracle'' setting, where the gold graph structure is provided.
\noindent\textbf{Attribute F1:} Following the original description of semantic proto-roles as binary attributes \citep{dowty.d.1991},
we also measure whether the direction of the attributes matches that of the gold annotations, e.g. whether a predicate is likely factual (\emph{factuality-factual} $>\theta$) or not (\emph{factuality-factual} $<\theta$).\footnote{Like $\rho$, we use an oracle decode of the structure here.} 
We tune $\theta$ per attribute type on validation data.
It is abbreviated as F1 (attr), and along with $\rho$, measures performance on the attribute prediction task. 


\noindent\textbf{S-score:} 
Following the Smatch metric \citep{cai.s.2013}, which uses a hill-climbing approach to find an approximate graph matching between a reference and predicted graph, S-score \citep{zhang.s.2017} provides precision, recall, and F1 score for nodes, edges, and attributes. 
Note that while S-score enables us to match scalar attributes jointly with nodes and edges, for the sake of clarity we have chosen to bifurcate the evaluations: S-score for nodes and edges only (functionally equivalent to Smatch), and $\rho$ and F1 (attr) for attributes. 
We use two variants of S-score: one evaluates against full UDS arborescences with linearized syntactic subtrees included as children of semantic heads (abbreviated as F1 (syn)), while the semantics-only setting evaluates only on semantics nodes (F1 (sem)). 
This metric measures performance on the semantic graph structure prediction task. 

\section{Experiment 1: Results and Analysis}\label{sec:experiment1} 
\begin{figure}[h]
    \centering
    \vspace{-1.5em}
    \includegraphics[width=\linewidth]{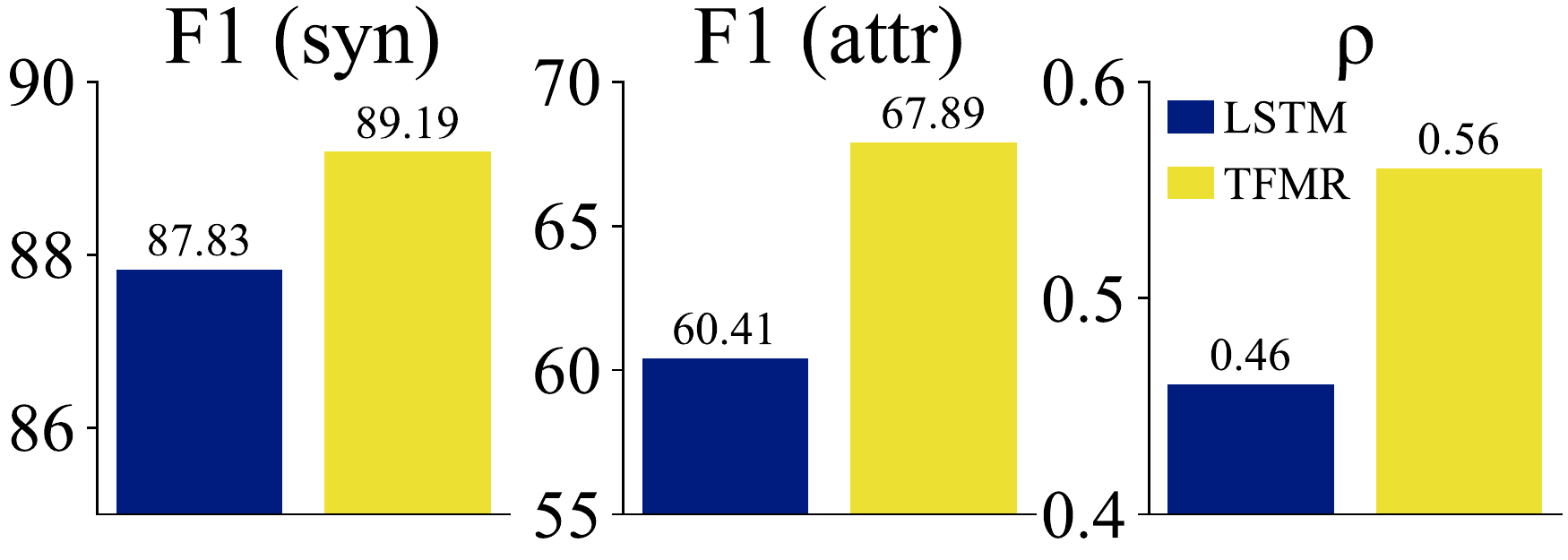}
    \vspace{-2em}
    \caption{Contrasting LSTM and TFMR performance on graph structure with syntax nodes as well as Pearson's $\rho$ and binarized attribute F1.}
    \label{fig:lstm_vs_tfmr_base}
    \vspace{-0.75em}
\end{figure}

\textbf{The Transformer outperforms the LSTM on UDS parsing.} We first observe that, with modifications and tuning, the Transformer architecture strictly outperforms the LSTM despite the relatively low number of training examples (12.5K). 
Fig.~\ref{fig:lstm_vs_tfmr_base}, which corresponds to Table~\ref{tab:big} rows 2 and 3, shows that the Transformer outperforms the LSTM on the S-score metric (with syntactic nodes included, following \citet{stengel-eskin.e.2020}) as well as attribute F1 and Pearson's $\rho$. Note that in this figure, as well as the others in this section, the vertical axis is scaled to highlight relevant contrasts.

\begin{figure}[h]
    \centering
    \vspace{-0.5em}
    \includegraphics[width=\linewidth]{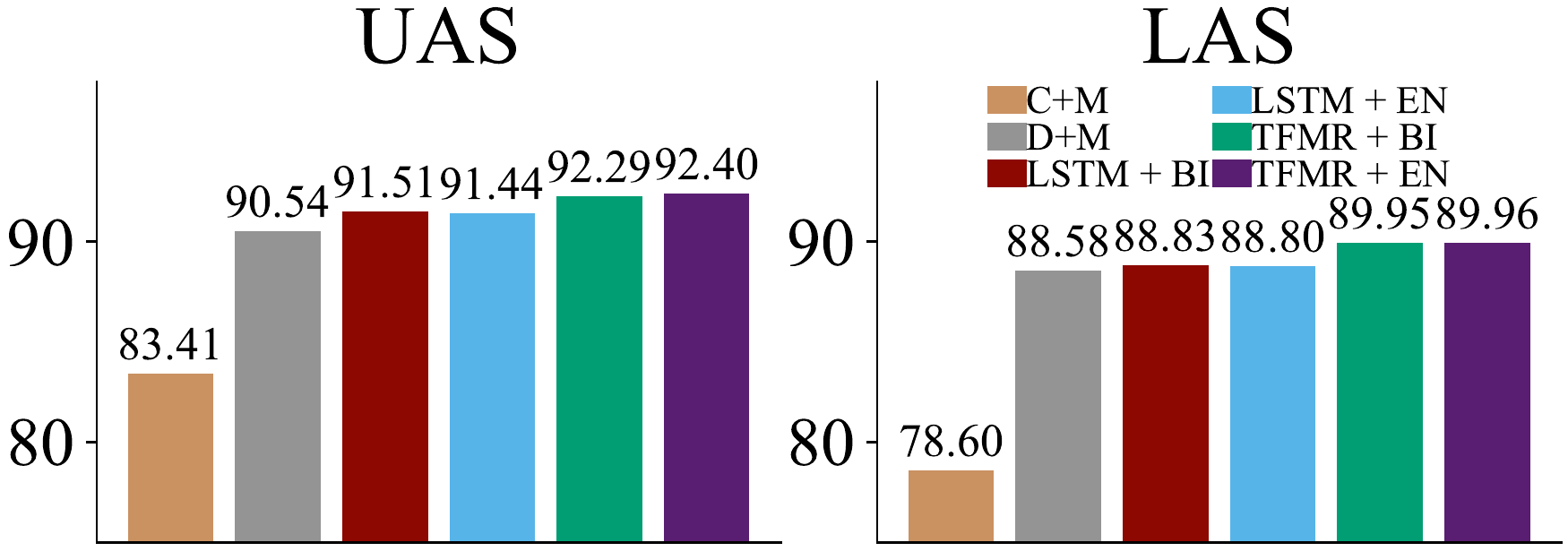}
    \vspace{-2em}
    \caption{LSTM and TFMR performance on English EWT UD parsing, contrasted with \citet{chen.d.2014} and \citet{dozat.t.2016} baselines. Models with semantic information (+ EN) outperform their syntax-only baselines (+ BI)} 
    \label{fig:lstm_vs_tfmr_syntax_only}
    \vspace{-1em}
\end{figure}

\textbf{Joint Transformer model slightly outperforms syntax-only model on syntactic parsing.} Fig.~\ref{fig:lstm_vs_tfmr_syntax_only}, corresponding to rows 4-6 and 11 in Table~\ref{tab:big}, shows that an LSTM encoder with a biaffine parser and no semantic decoder (LSTM + BI) outperforms both baselines \citep[C+M and D+M, respectively]{chen.d.2014, dozat.t.2016}.
Note that this model has no semantic signal, and is trained only on UD parsing.
In the LSTM case, the addition of the UDS semantic signal via the encoder-side model described in \S\ref{sec:syn_model} slightly lowers performance.
However, this is not the case for the Transformer; the syntax-only Transformer (TFMR + BI) model outperforms the LSTM model, and is slightly outperformed by the joint syntax-semantics Transformer model. 

\begin{figure}[h]
    \centering
    \vspace{-0.5em}
    \includegraphics[width=\linewidth]{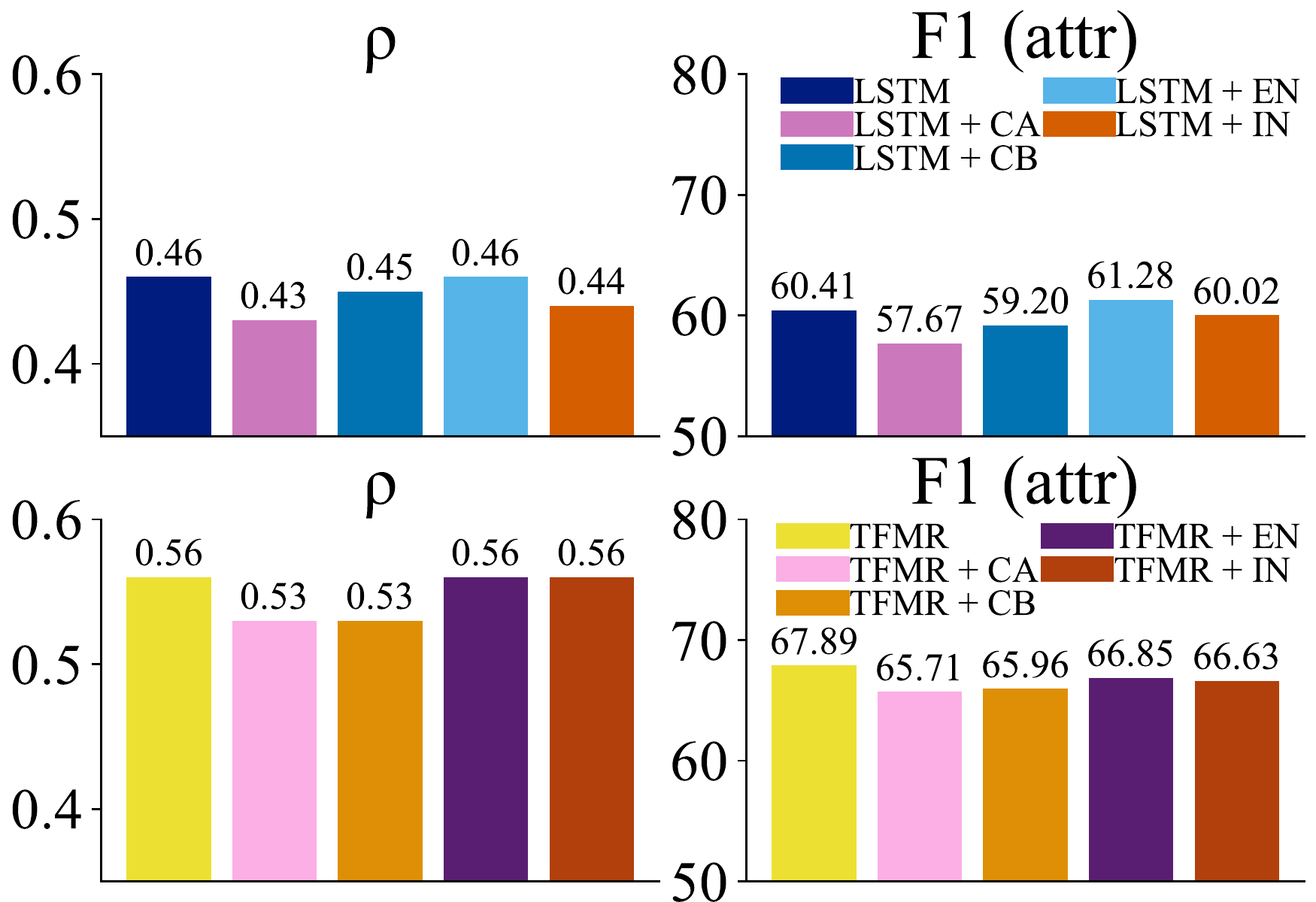}
    \vspace{-2em}
    \caption{LSTM and TFMR attribute $\rho$ and binarized F1 score.} 
    \label{fig:lstm_tfmr_sem_attr}
    \vspace{-1em}
\end{figure}

\textbf{Joint training has little impact on attribute metrics for non-baseline models.} Fig.~\ref{fig:lstm_tfmr_sem_attr} shows Pearson's $\rho$ and binarized attribute F1 (with $\theta$ tuned on the development set); this corresponds to the $2^\text{nd}$ two rows and final 10 rows of Table~\ref{tab:big}. 

We see that both for the LSTM and the Transformer, the encoder-side model has about the same performance for $\rho$ and attribute F1 as the UDS-only model, and the Transformer variants consistently out-perform their LSTM counterparts. 
In contrast, the addition of syntactic information through concatenation (concat-before, concat-after) seems to diminish the performance on these metrics.
For the LSTM, the intermediate model has lower performance than the encoder-side variant, while for the Transformer it is almost identical.

\begin{figure}[h]
    \centering
    \vspace{-0.5em}
    \includegraphics[width=\linewidth]{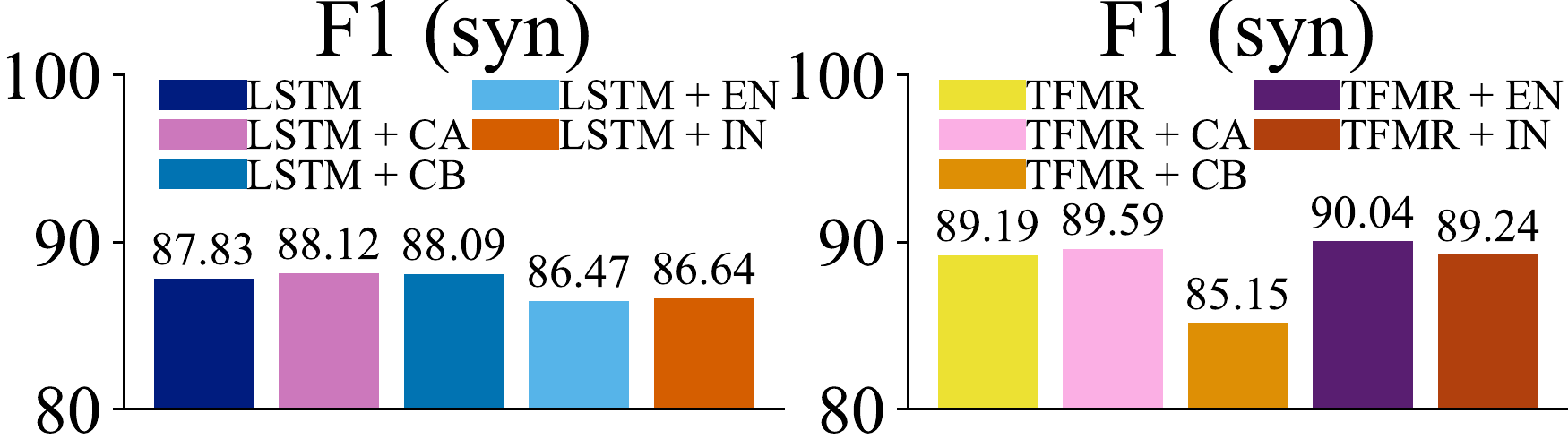}
    \vspace{-2em}
    \caption{LSTM and TFMR S-score F1 (with syntax nodes included).} 
    \label{fig:lstm_tfmr_sem_structure}
    \vspace{-1em}
\end{figure}

\textbf{Joint parsing slightly improves semantic structural performance.} Fig.~\ref{fig:lstm_tfmr_sem_structure} shows the structural F1 computed by S score, where we observe that the LSTM's performance, which is lower than the Transformer's in the baseline setting, benefits most from the concatenation settings, while suffering under the encoder and intermediate settings. 

By way of contrast, the Transformer, whose baseline performance is higher, benefits most from the encoder-side biaffine parsing setting, which also boasts the best UD performance (cf. Fig.~\ref{fig:lstm_vs_tfmr_syntax_only}). 

\begin{table*}[h]
\centering
\resizebox{\textwidth}{!}{%
   \begin{tabular}{lcccccccccc}
    \hline
    Model & P (syn) & R (syn) & F1 (syn) &  P (sem) & R (sem) & F1 (sem) & Attr. $\rho$ & Attr. F1 & UAS & LAS \\
    \hline
    SE 2020 & 84.97 & 78.52 & 81.62 & 91.28 & 87.23 & 89.21 & 0.34 & 50.66 & --- & ---\\
    LSTM & 89.90 & 85.85 & 87.83 & 89.24 & 87.47 & 88.34 & 0.46 & 60.41 & --- & ---\\
    TFMR & 90.04 & 87.98 & 89.19 & 92.26 & 91.09 & 91.67 & 0.56 & 67.89 & --- & ---\\
    C+M & 84.83 & 75.22 & 79.74 & 84.72 & 88.51 & 86.57 & --- & --- & 83.41 & 78.60\\
    D+M & --- & --- & --- & --- & --- & --- & --- & --- & 90.54 & 88.58\\
    LSTM + BI & --- & --- & --- & --- & --- & --- & --- & --- & 91.51 & 88.83\\
    LSTM + CB & 88.16 & 88.01 & 88.09 & 92.26 & 90.67 & 91.46 & 0.45 & 59.20 & 54.44 & 52.75\\
    LSTM + CA & 88.58 & 87.67 & 88.12 & 92.36 & 91.30 & 91.83 & 0.43 & 57.67 & 50.89 & 49.33\\
    LSTM + EN & 87.44 & 86.47 & 86.47 & 92.52 & 90.90 & 91.70 & 0.46 & 61.28 & 91.44 & 88.80\\
    LSTM + IN & 86.80 & 86.49 & 86.64 & 91.50 & 90.35 & 90.92 & 0.44 & 60.02 & 91.00 & 88.31\\
    TFMR + BI & --- & --- & --- & --- & --- & --- & --- & --- & 92.29 & 89.95\\
    TFMR + CB & 92.87 & 78.62 & 85.15 & 93.66 & 85.42 & 89.35 & 0.53 & 65.96 & 58.05 & 56.75\\
    TFMR + CA & 91.31 & 87.94 & 89.59 & 93.15 & 91.94 & 92.54 & 0.53 & 65.71 & 51.13 & 50.07\\
    TFMR + EN & 91.09 & 89.01 & 90.04 & 93.76 & 91.50 & 92.61 & 0.56 & 66.85 & 92.40 & 89.96\\
    TFMR + IN & 91.50 & 87.09 & 89.24 & 93.26 & 91.25 & 92.24 & 0.56 & 66.63 & 92.16 & 89.52\\
\end{tabular}
}
\vspace{-1em}
\caption{Syntactic and semantic metrics across all models. Note that binarized semantic attribute F1 and $\rho$ are computed w.r.t. to the models trained with linearized syntactic yields, while the (sem) S-score metrics are reported on models trained on semantics nodes alone (on the decoder side).} 
\label{tab:big}
\vspace{-1.5em}
\end{table*}

While the concat-after setting offers S score improvements for both encoder/decoder types, the syntactic performance in this setting is very poor ($<60$ UAS).
The Transformer encoder-side multitask model is able to improve structural performance for the encoder-side while \emph{simultaneously} boosting UAS and LAS (see Fig.~\ref{fig:lstm_vs_tfmr_syntax_only}). 

These results demonstrate that explicitly incorporating a syntactic signal into a transductive semantic parsing model can be done without damaging semantic performance, 
both for UDS attributes and structure. 
Perhaps more surprisingly, the semantic signal coming from the UDS attributes and structure improves the syntactic performance of the model when the syntactic model is able to take advantage of the lexicalized nature of UD. 
Note that due to the parallel nature of the UD and UDS data, we can conclude that the improvements here result from the additional structural signal, and not merely from the addition of more sentences.  
We see that for the concatenation settings, while the semantic structural performance may increase, the syntactic parsing results are dismal. This is true whether we concatenate the syntactic graph before or after the semantic one.  
This may be explained by the fact that, by using a transductive model for a lexicalized parsing task like UDS, we are complicating what the model needs to learn. 
Rather than simply labelling existing nodes, the model must reproduce these nodes via source-side copying. 

While \emph{prima facie}, we would expect the intermediate model to outperform the multitask encoder-side model, as its decoder has explicit access to the syntactic parse, we see that this is not the case; it shows lower structural and attribute performance.   
This represents a direction for future work. 

\begin{figure} [h]
    \centering
    \vspace{-1em}
    \includegraphics[width=\linewidth]{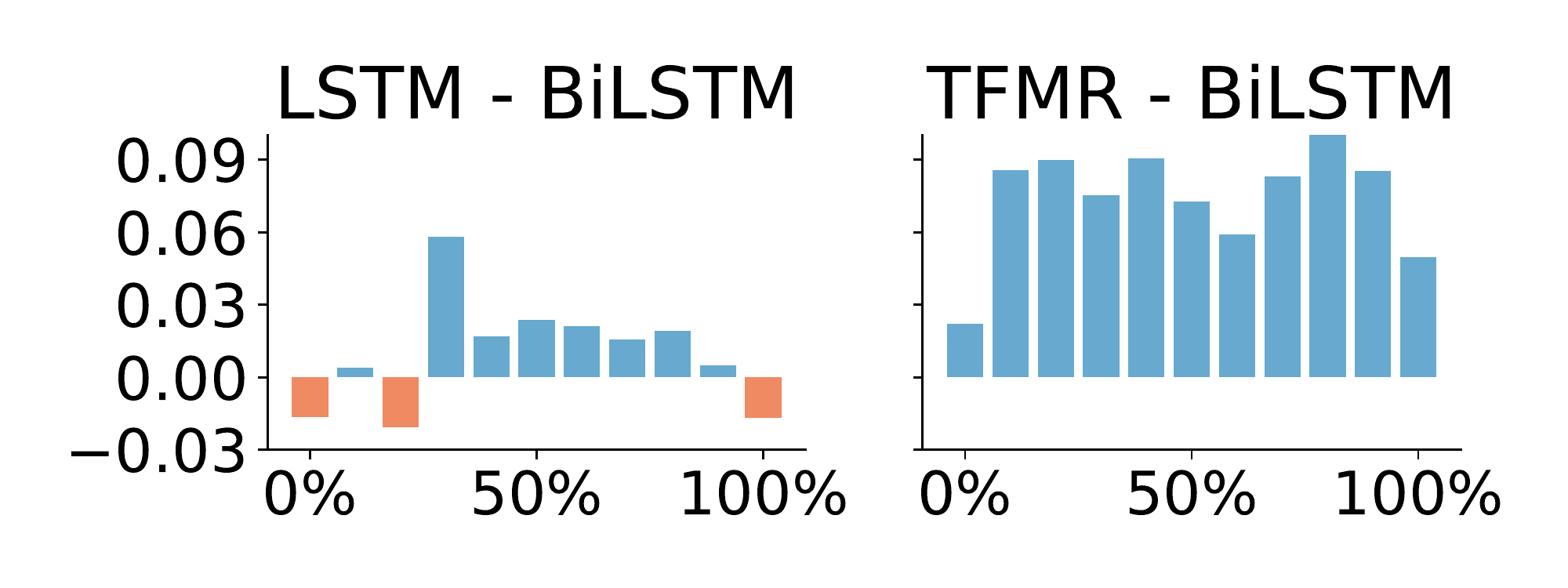}
    \vspace{-2.75em}
    \caption{$\Delta$ in mean $\rho$ for each sentence position between (left) LSTM and BiLSTM and between (right) TFMR and BiLSTM. The TFMR's gains in $\rho$ come mainly from the middle percentiles.}
    \label{fig:lstm_tfmr_pearson_percentiles}
    \vspace{-1.5em}
\end{figure}

\vspace{-0.5em}
\paragraph{The Role of the Encoder} The results in Fig.~\ref{fig:lstm_tfmr_sem_attr} show that the Transformer-based model has a heavy advantage over the LSTM-based model in terms of attribute prediction. 
This might be due to an improved ability by the Transformer encoder to incorporate signals from across the input, since the self-attention mechanism has equal access to all positions, while the BiLSTM has only sequential access which may become corrupted or washed out over longer distances.
Given the highly contextual nature of UDS inferences, we would expect a model which better captures context to have a distinct advantage, as the crucial tokens for correctly inferring an attribute value may be found in a distant part of the input sentence: for example, if we wish to infer the factuality of ``\emph{left}'' in the sentence ``Bill eventually confessed to the officers that, contrary to his previous statements, Joan had \emph{left} the party early,'' most of the signal would come from the token ``confessed,'' producing a high score. 
The construction of UDS and its attributes' scalar nature allow us to test this hypothesis by examining the Pearson correlation between predicted and true attributes at different positions in the input.\footnote{We use the syntactic head nodes of each semantic node to propagate positional information to the semantic nodes.} 
In order to compare the correlations across sentences, we group the predicted and reference attributes into 10 percentile ranges, based on the ratio of the node position and the sentence length, i.e. what percentage into the sentence the node occurs. 
We then average all Pearson $\rho$ values across all attribute types in each bin, obtaining average $\rho$ values by sentence completion percentile. 
With this data, we can compare two models on a very fine-grained level, asking questions such as, ``how much better does the Transformer model do than the BiLSTM on nodes that are between 0\% and 10\% into the sentence.''
Fig.~\ref{fig:lstm_tfmr_pearson_percentiles} shows such comparisons between a unidirectional left-to-right LSTM encoder and the bidirectional LSTM, and between the bidirectional LSTM and the Transformer.
While the unidirectional LSTM actually outperforms the BiLSTM in the central percentiles, it struggles near the edges. 
This could be explained on the left edge by a lack of right context, and on the right edge by difficulties with long-range dependencies. 
Furthermore, the Transformer outperforms the BiLSTM at all positional percentiles, but particularly in the central regions, suggesting that, while the BiLSTM is able to incorporate contextual information well at the edges of a sentence, the information is diluted in the central region, while the Transformer's self-attention mechanism is equally able to draw from arbitrary positions at all timesteps. 

\section{Experiment 2: Tuning BERT}\label{sec:tuning}
The results in Fig.~\ref{fig:lstm_vs_tfmr_syntax_only} and Fig.~\ref{fig:lstm_tfmr_sem_structure} not only demonstrate that the addition of one structural modality (i.e. syntax, semantics) can benefit the other, but also suggest that these signals are complementary to the signal already given by the input features, which include contextualized features obtained from BERT. 
This stands in contrast to previous results by \citet{swayamdipta.s.2019} that the benefits to be gained from multitask learning with shallow syntactic objectives are largely eclipsed by contextualized encoders. However, we note that our models require full UD parses in the multitask settings rather than a light scaffolding. 

If indeed the combination of UDS and UD signals provides information not yet encoded in BERT, then fine-tuning BERT with these signals should yield additional benefits.
Following observations that syntax and semantics are encoded to varying degrees at different depths in contextualized encoders \citep{hewitt.j.2019, tenney.i.2019, jawahar.g.2019} with syntactic information typically lower in the network, we explore the trade-off between freezing and tuning various layers of the BERT encoder. 
Specifically, we tune the top $n$ layers, starting from a completely frozen encoder and moving to tuning all 12 layers.\footnote{As the BERT encoder is pre-trained, a separate static learn rate of $1e-5$ was used to optimize the BERT parameters.} 

\begin{figure}
    \centering
    \includegraphics[width=\columnwidth]{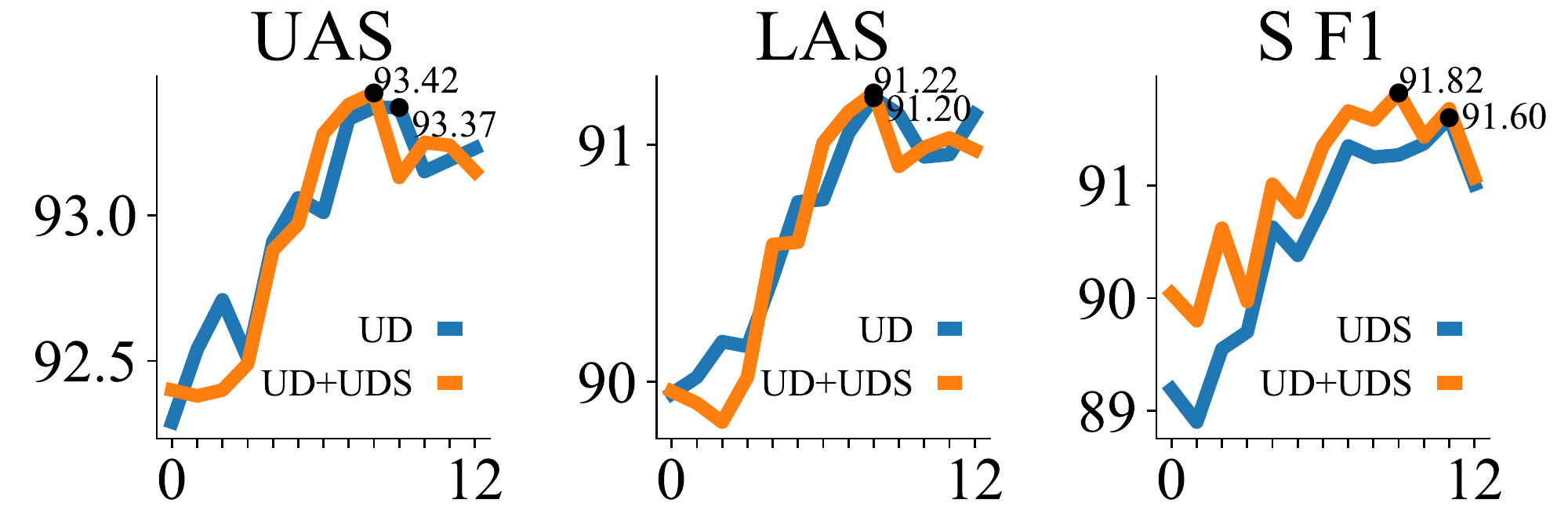}
    \vspace{-2em}
    \caption{Test metrics when freezing/tuning different levels of BERT. X-axis represents number of layers tuned (from the top layer). Tuning the encoder provides significant benefits over a frozen encoder (0 layers tuned) but the optimal number of tuned layers is not the full 12 layers.}
    \label{fig:tuning}
    \vspace{-1.5em}
\end{figure}

Intuitively, one might expect to see a monotonic increase as the number of tuned layers increases, as each additional unfrozen layer provides the model with more capacity. 
However, the results presented in Fig.~\ref{fig:tuning} show a more nuanced trend: while the performance across syntactic and semantic metrics increases up to a point, they begin to decrease again when additional layers are unfrozen. 
This may be due to data sparsity; given the relatively small size of the UDS corpus, the addition of too many parameters may encourage overfitting, resulting in decreased test performance. 
Note that the encoder-side model all three panels of Fig.~\ref{fig:tuning} is the same model, i.e. the best UAS, LAS, and S score performance is obtained by the same model, and that the performance of the joint model at any given tuning depth typically falls above that of the baseline.  

In contrast to the findings of \citet{glavas.g.2020}, who conclude that the benefits of UD pre-training for semantic language understanding tasks are limited when using contextualized encoders, our results in \S\ref{sec:experiment1} show a small but consistent positive effect of syntactic information on semantic parsing, as well as improved syntactic performance from a semantic signal. 
Furthermore, our results here show that the UD signal can actually be used to fine-tune a contextualized encoder, which benefits not only the UD parsing performance but also the UDS performance. 
In fact, after training and evaluating their model (which, to our knowledge, has the highest performance to date on EWT) on our cleaned subset of EWT, we find that our best performing UAS/LAS values, 93.42 and 91.22, outperform their values of 92.83 and 90.11. 
These values also slightly outperform the syntax-only version of the same model, with the same amount of tuning. 
The tuned encoder-side model also provides the best semantic performance, with a max score of 91.82, compared to 90.04 in the TFMR + EN setting (cf. Table~\ref{tab:big}). 

\vspace{-0.8em}
\paragraph{Prepositional Phrase Attachment Ambiguity} Looking at the results in Figs.~\ref{fig:lstm_vs_tfmr_syntax_only} and \ref{fig:lstm_tfmr_sem_structure}, which are mirrored in the tuned results visualized in Fig.~\ref{fig:tuning}, a natural question to ask is where the syntactic performance gains are coming from in the encoder-side model. 
One hypothesis, in line with literature on semantic bootstrapping, is that the semantic signal helps the model to discriminate between ambiguous parses. 
Consider, for example, the sentence ``I shot an elephant in my pyjamas.'' 
Syntactically, there are two valid heads for the prepositional phrase (PP) ``in my pyjamas'', but the semantics of the phrase indicates to us that it is less likely to be attached to ``elephant''. 
Perhaps adding an explicit semantic signal, like that of UDS, would improve a syntactic parser's ability to disambiguate sentences like this. 
In order to test this hypothesis, we use a dataset introduced by \citet{gardner.m.2020}, consisting of 300 sentences with UD annotations. 
150 sentences were chosen from a combination of English UD treebanks with potential PP attachment ambiguities, 75 with PP with nominal heads, and 75 with verbal heads. 
Minimal semantic changes were then made to the sentence to switch the head (i.e. nominal heads were switched to verbal heads). 
For example, the sentence, ``They demanded talks \emph{with local US commanders}'' becomes: ``They demanded talks \emph{with great urgency}'' (noun to verb). 

\begin{figure}
    \centering
    \vspace{-1em}
    \includegraphics[width=\columnwidth]{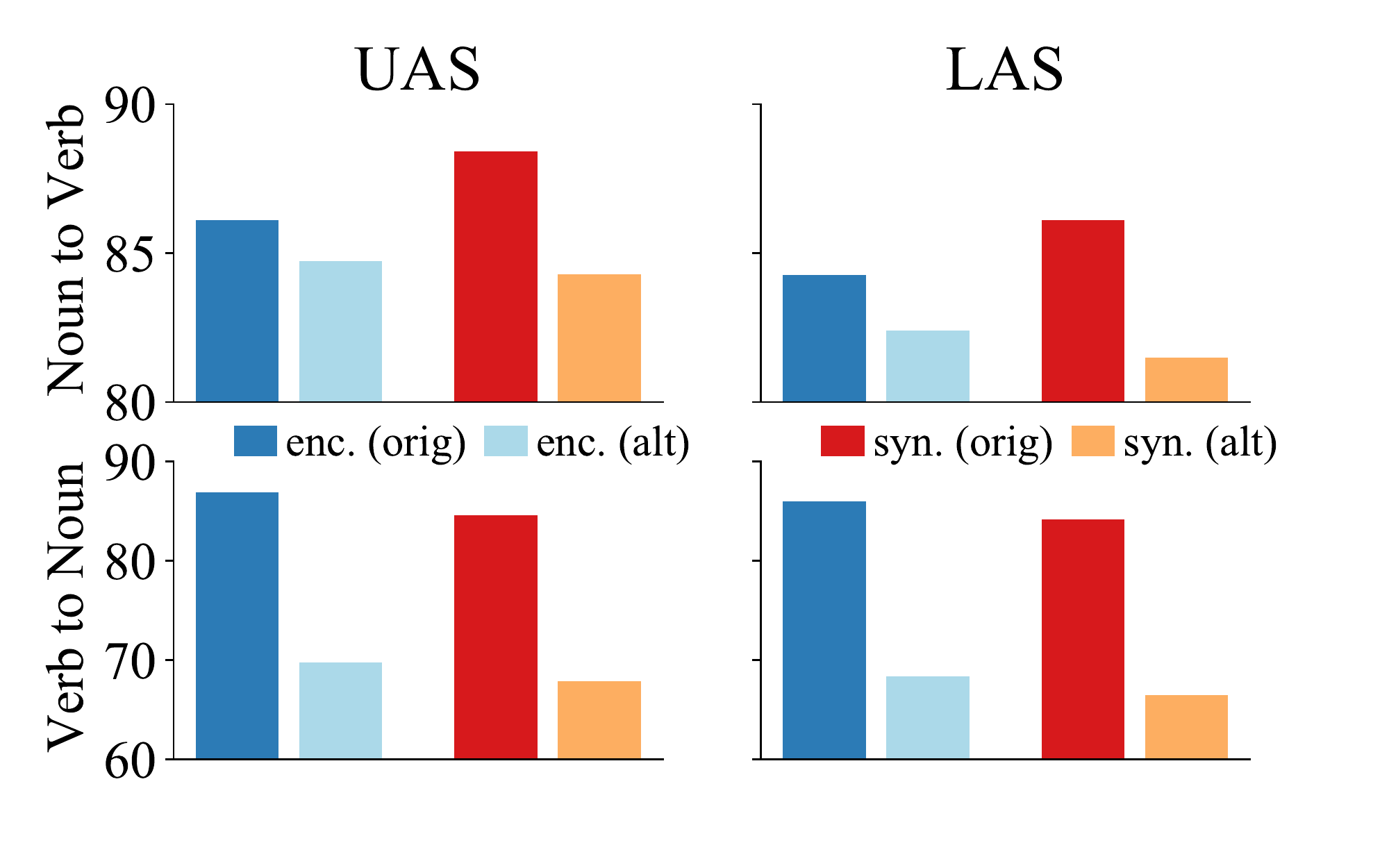}
    \vspace{-3em}
    \caption{Encoder-side and syntax-only performance on sentences with PP attachment ambiguities. A joint syntax-semantics model is slightly more robust to manual adjustment of the prepositional head than a syntax-only model.}
    \label{fig:pp_attach}
    \vspace{-1.5em}
\end{figure}

A model's performance on this dataset is measured not only it its raw performance on the unaltered sentences, but, crucially, by its performance on the altered ones. 
As the altered sentences are constructed to be different from those seen in the training set, we expect there to be a drop if the model has learned simple heuristics (e.g. always attach to a noun) rather than robust rules based on semantic understanding. 
An ideal model would have high performance and no drop. 
Fig.~\ref{fig:pp_attach} compares the tuned syntax-only UD baseline (right column) and the tuned UDS parser with encoder-side parsing (left column) on this task, both for noun-to-verb and verb-to-noun alterations. 
In all cases, we see a significant drop in performance on the altered examples. 
In the noun-to-verb case, we see that, while the syntax-only baseline's initial performance is higher, it experiences a larger drop in performance than the encoder-side model, with its performance on the altered dataset being lower on UAS and LAS. 
In the verb-to-noun case, while both models undergo roughly the same major performance loss in the altered context, the initial performance of the encoder-side model is higher. 

\begin{figure}[h]
    \centering
    \includegraphics[width=\columnwidth]{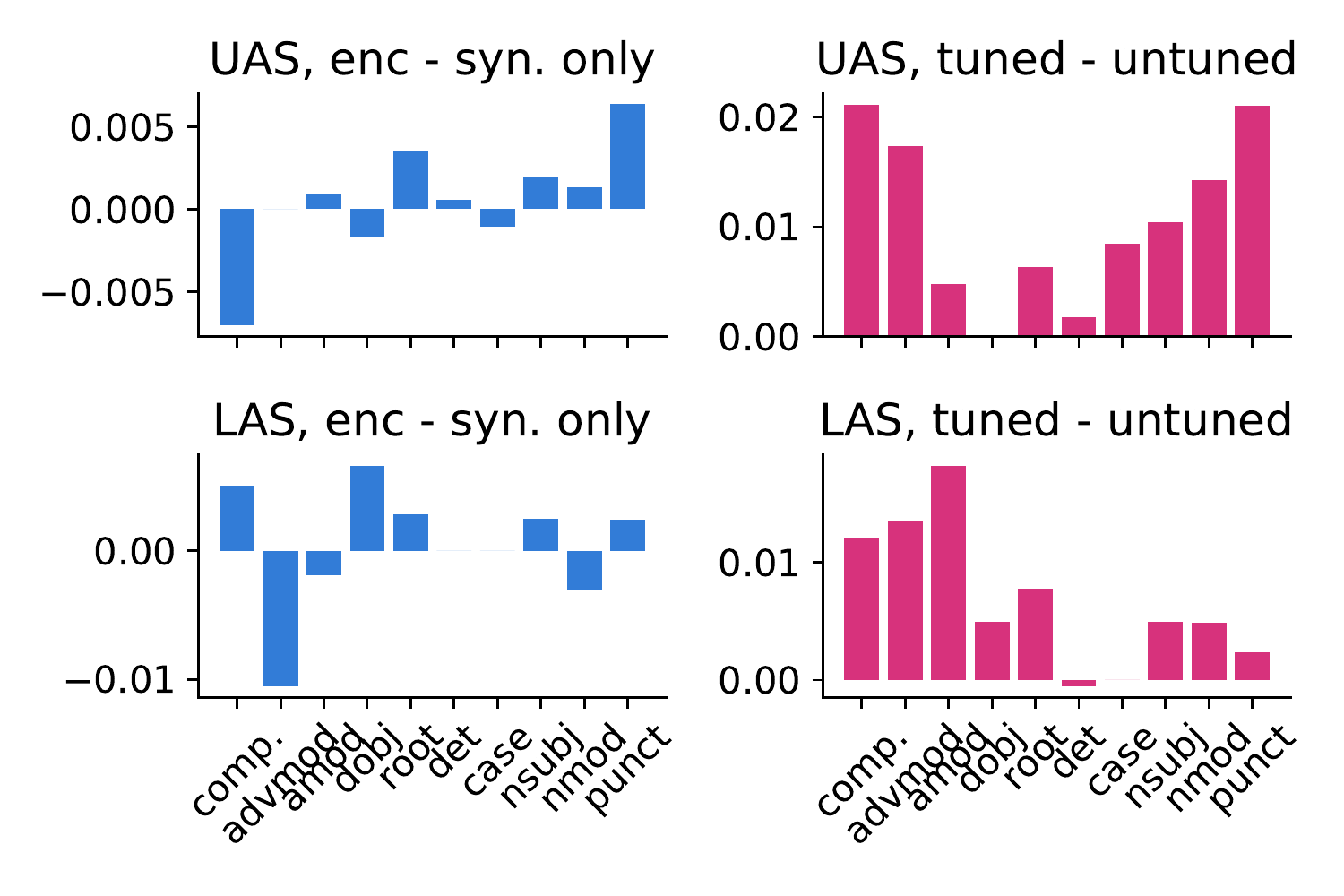}
    \vspace{-2.5em}
    \caption{$\Delta$ between (left) a joint UDS-UD model and a UD only model and (right) a joint UDS-UD model with an encoder tuned through layer 4 vs. one with a frozen encoder.}
    \label{fig:uas_comparison}
    \vspace{-1em}
\end{figure}

These results, taken together, suggest that the addition of the UDS signal may provide a small benefit when disambiguating PP attachment ambiguities. 
However, such ambiguities are fairly rare in UD corpora, and are thus unlikely to explain the whole difference between the models. 
To this end, we examine the difference in UAS performance between systems on the 10 most frequent relation types in Fig.~\ref{fig:uas_comparison}.
When comparing a joint UD-UDS parser, we see that small gains are realized for the most frequent relations, but some relations suffer minor losses as well. 
In contrast, when comparing the tuned and untuned systems, nearly all the most frequent relations see fairly large improvements. 

\vspace{-0.6em}
\paragraph{UDS attributes and UD relations} The close link between UD parses and the UDS annotations in the dataset allows us not only to train multitask models for joint syntactic and semantic parsing, but also to inspect the interactions between syntactic relations and semantic attributes more closely. Each cell in Fig.~\ref{fig:heatmap} shows the Pearson $\rho$ between true and predicted attributes for a variety of UDS annotations conditioned on UD dependency relations. 
The node attributes (annotated on semantic nodes in the UDS graph) are paired with the UD relation of the corresponding syntactic head node. 
Predictions are obtained from the best tuned model with encoder-side UD parsing (TFMR + EN), under an oracle decode of the graph structure. 

We see variation across a given attribute and dependency relation. 
For example, factuality annotations display a high $\rho$ value for {\tt{root}} and {\tt{conj}} annotations, but a lower correlation for {\tt{xcomp}}. 
\begin{figure*}
    \centering
    \vspace{-0.5em}
    \includegraphics[width=\linewidth]{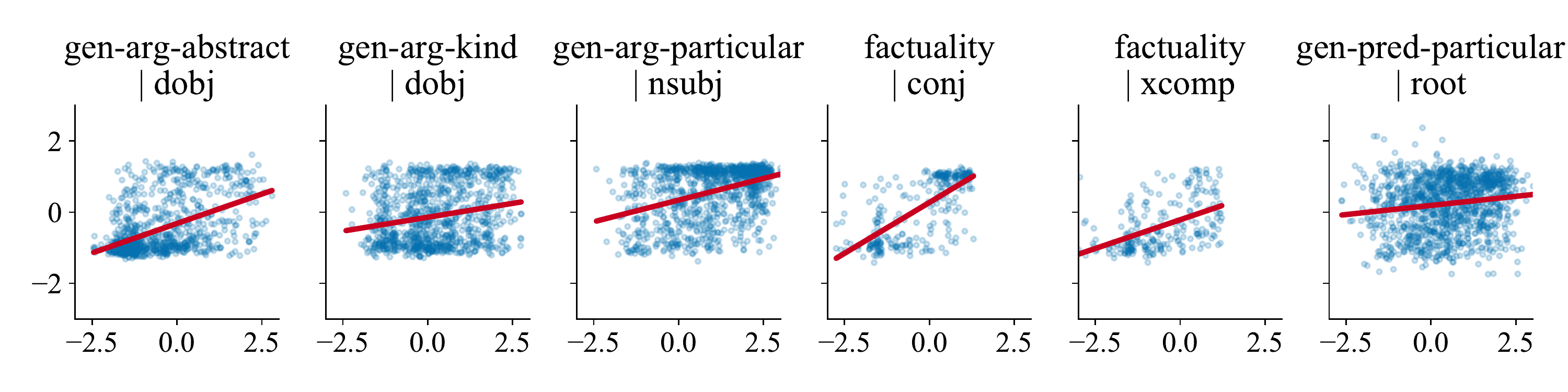}
    \vspace{-2.5em}
    \caption{True and predicted values for UDS attributes at UD relation types, outlined in Fig.~\ref{fig:heatmap}}
    \vspace{-1.5em}
    \label{fig:correlations}
\end{figure*}
These correlations are visualized in Fig.~\ref{fig:correlations}, where we plot the true vs. predicted value, with the line defined by $\rho$ overlaid. 
The close correspondence between UD and UDS lets us observe this type of discrepancy, which echoes findings by \citet{white.a.2018}, who used factuality prediction to probe neural models' ability to make inferences based on lexical and syntactic factuality triggers. 
Furthermore, it is in holding with semantic theories, as the {\tt{xcomp}} relation is used for open clausal complements, i.e. non-finite embedded clauses, with an overt control subject in the main clause (e.g. object or subject control).
In English, {\tt{xcomp}} relations correspond to infinitival embedded clauses, e.g. ``I remembered \emph{to turn off the stove}.'' 
As pointed out by \citet{white.a.2020b}, factuality inferences are particularly hard in these contexts, as they are not only sensitive to the lexical category of the embedding predicate (i.e. ``remembered'' vs. ``forgot'') but also its polarity (i.e. ``remembered'' vs. ``didn't remember''). 
This separates them from finite clausal complements, where a matrix negation still results in the same factuality inference; e.g. in both ``I remembered \emph{that I turned off the stove}'' and ``I didn't remember \emph{that I turned off the stove}''  we infer that the stove was turned off. 
Furthermore, {\tt{xcomp}} relations are present in object and subject control cases, which may be difficult even for human speakers to acquire \citep{chomsky.c.1969, cromer.r.1970}.

Beyond comparing our model predictions to theoretical predictions at the syntax-semantics interface, we can also use this analysis to examine the data on which the model was trained. 
For instance, homing in on the {\tt{genericity-arg-kind}} annotations (reflecting the level to which an argument refers to a \emph{kind} of thing) for direct objects {\tt{dobj}}, we see that for some examples, while the model prediction differs from the annotation, it is not wrong per se. 
One example is, ``Take a \emph{look} at this spreadsheet'' where ``look'' is annotated as high for {\tt{kind}} (1.41), but predicted as low (-1.09). 
In another example, ``...I could not find one \emph{place} in Tampa Bay that sells...'', the argument ``place'' has a high predicted {\tt{kind}} value (0.72) but is annotated otherwise (-0.87). 
In both these cases, one could argue that the model's prediction is not entirely incorrect.

\begin{figure}[H]
\centering
\vspace{-1.8em}
    \begin{tabular}{l}
    \vspace{-3.5em}
        \includegraphics[width=\linewidth]{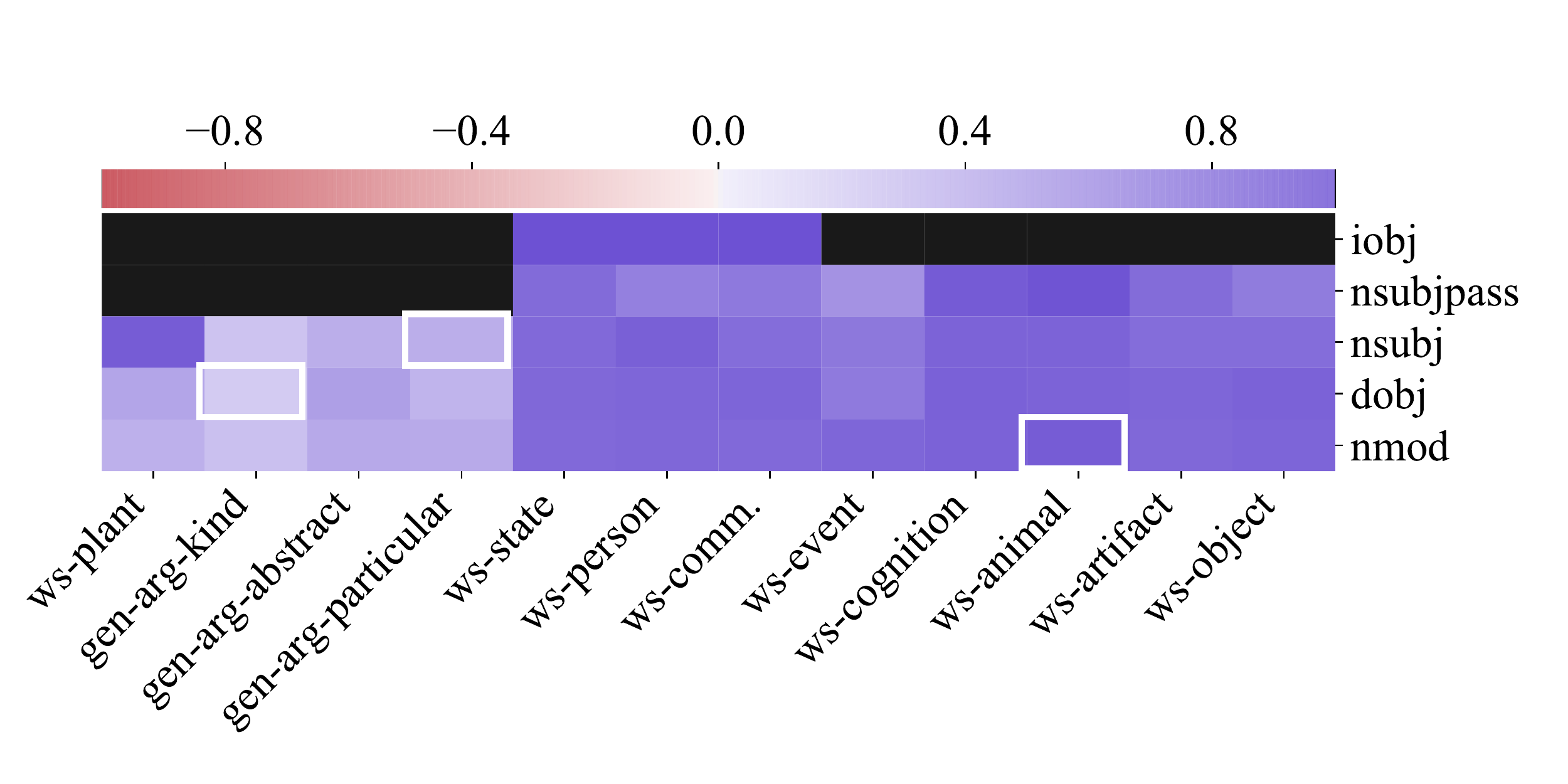}\\
        \vspace{-1.5em}
        \includegraphics[width=\linewidth]{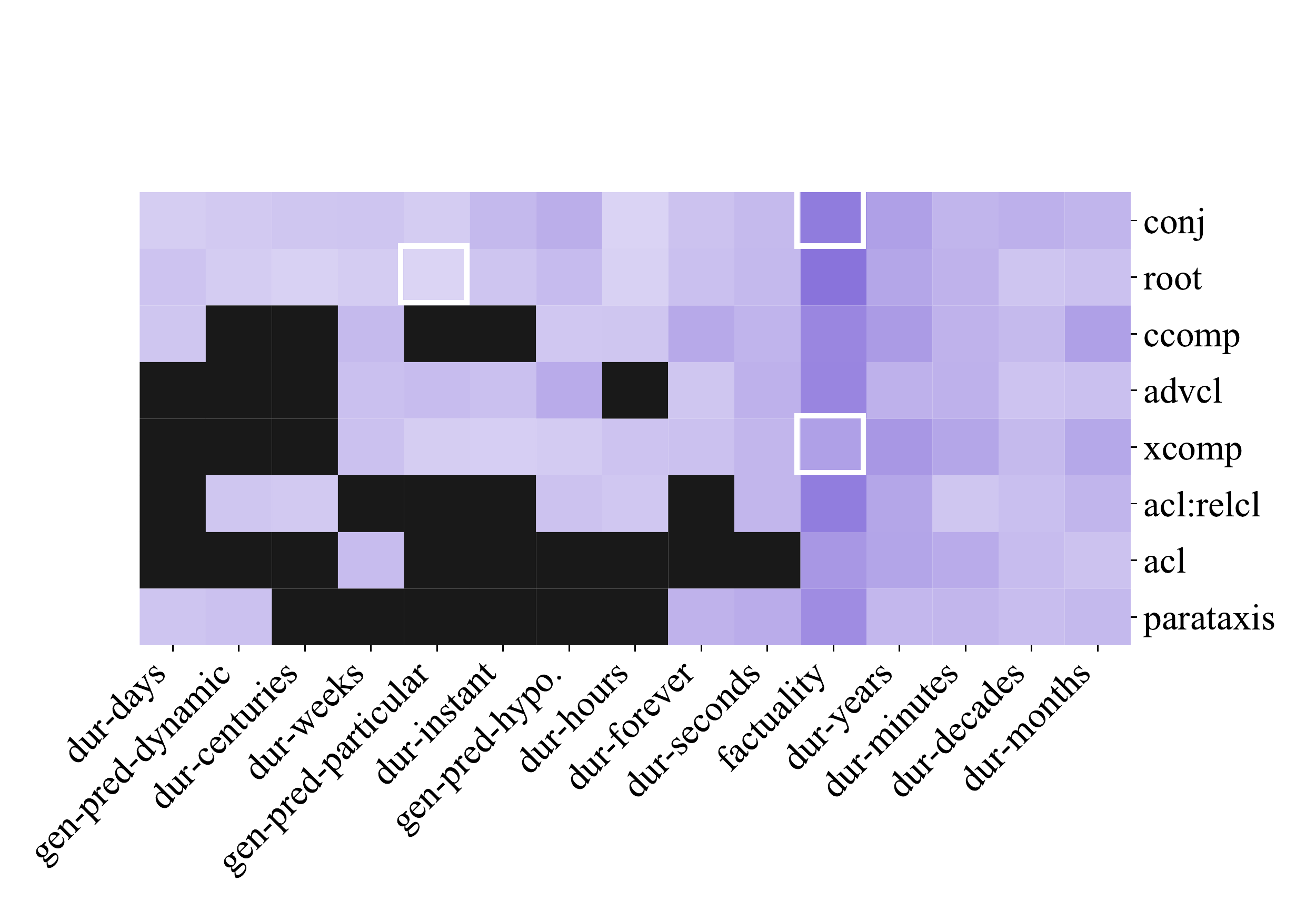}
    \end{tabular}
        \caption{$\rho$ on argument (top) and predicate (bottom) UDS properties (subset) at relevant UD relations. Black cells indicate no significant correlation. Outlined boxes plotted in detail in Fig.~\ref{fig:correlations}.}
        \label{fig:heatmap}
        \vspace{-1em}
\end{figure}  

\section{Experiment 3: Multilingual Parsing}\label{sec:multiling} The results in \S\ref{sec:experiment1} that English UD parsing and UDS parsing are mutually beneficial naturally give rise to a follow-up question: does this relationship extend to a multilingual setting. 
As in the monolingual case, we explore both the impact of UD parsing on UDS, and vice versa. 
UD is by design highly multilingual, spanning scores of languages from a diverse typological range, often with multiple treebanks per language. 
This has led to interest in evaluating the performance of UD parsing models not just on English, but across a range of languages and language families; both the 2017 and 2018 CoNLL shared tasks focused on multilingual UD parsing \citep{zeman.d.2017, zeman.d.2018}. 
The introduction of multilingual contextualized encoders, such as mBERT \citep{devlin.j.2019,devlin.j.2018} and XLM-R \citep{conneau.a.2019} has enabled models to perform UD parsing in multiple languages simultaneously by using features obtained by a single multilingual encoder \citep{schuster.t.2019, kondratyuk.d.2019}.
   
By initializing weights for one task (syntactic or semantic) with weights learned on the other, and leveraging the shared input representation space of XLM-R, we examine bottom-up effects of representations learned with a syntactic objective on semantic parsing performance, and top-down effects the semantic objective on syntactic performance. 
Note that unlike in \S\ref{sec:experiment1}, we do not have parallel data in these settings, leading to the use of pre-training rather than simultaneous multitask learning.
Note also that we are examining the relationship between \emph{English} semantic parsing and multilingual, \emph{non-English} syntactic parsing. 
We do not make use of pre-trained type-level word embeddings in these experiments, leading us to expect slightly lower absolute performance on the UDS parsing task as compared to Table~\ref{tab:big}.
Based on the syntactic results in \S\ref{sec:experiment1}, we explore only the Transformer models in our multilingual experiments. We tune the XLM-R encoder through layer 5, based on our observations on the development set in \S\ref{sec:experiment1}.\footnote{We also re-tuned the Transformer hyperparameters, as the input space changed from BERT and GloVe to XLM-R.}

\vspace{-0.8em}
\paragraph{Languages}8 languages in 5 families from the 2018 CoNLL Shared Task \citep{zeman.d.2018} were chosen, across both higher and lower resource settings. Table~\ref{tab:langs} gives further details and highlights the range of resource settings examined.\footnote{Note that despite its relatively large test set size, the Kazakh train set is very small ($n=31$). Of these sentences, 3 were used for validation, leaving 28 train sentences.} 

\begin{table}[]
    \centering
    \begin{tabular}{lllll}
    \hline
         Code & Language & Family & \#Train  & \#Test \\
    \hline
         AF & Afrikaans & Germanic & 2630 & 850 \\
         DE & German & Germanic &  13814 & 977 \\
         FI & Finnish & Uralic &  27198 & 4422 \\
         FR & French & Romance & 17836 & 1598 \\
         GL & Galician & Romance & 2272 & 861 \\
         HU & Hungarian & Uralic & 910 & 449 \\
         HY & Armenian & Armenian & 1975 & 278 \\
         KK & Kazakh & Turkic & 28  & 1047 \\
    \hline
    \end{tabular}
    \vspace{-1em}
    \caption{Language data and \# train/test sentences} 
    \label{tab:langs}
    \vspace{-1.5em}
\end{table}

\vspace{-0.8em}
\paragraph{Bottom-up Effects} 
Examining bottom-up effects of syntax on semantics, we pre-train a multilingual UD model on all 8 languages simultaneously, using alternating batches from each language and capping each epoch at 20,000 examples. 
We then initialize the encoder and biaffine parser in UDS parser with the weights learned on UD parsing and continue training on joint UDS and EWT UD parsing, as in Experiment 1. 
In Table~\ref{tab:bottom-up} we see that syntactic parsing performance on English EWT UD parsing improves with pretraining for the intermediate model, but decreases for the encoder-side model. 
A similar trend holds for the S-score, where the intermediate model improves with pre-training while the encoder-side model's performance suffers. 
However, for attribute F1 and $\rho$, the encoder-side model with pre-training outperforms the encoder-side model without, while the opposite is true of the intermediate model, whose performance decreases with pretraining. 
Unlike in \S\ref{sec:experiment1}, the intermediate model outperforms the encoder-side model here, obtaining the highest overall UD parsing score of any model looked at thus far. 

The UAS/LAS of the pre-trained intermediate model is the strongest even when compared against the best monolingual models in Table~\ref{tab:big}. In fact, when trained and evaluated on the entirety of the EWT UD corpus, the XLM-R-based intermediate model with pre-training obtains the same exact UAS/LAS performance the best XLM-R model reported by \citet{glavas.g.2020}: 93.1 UAS, 90.5 LAS. 
Since our model is additionally capable of performing UDS parsing at a level competitive with the best system presented in Table~\ref{tab:big}, we encourage others to make use of it in the future. 


\begin{table}[h]
        \centering
        \resizebox{\linewidth}{!}{%
        \begin{tabular}{lccccccc}
            \hline
            Model & P & R & F1 & Attr. $\rho$ & Attr. F1 & UAS & LAS \\
            \hline

            TFMR & 90.80 & 89.08 & 89.93 & \textbf{0.51} & \textbf{63.62} & --- & ---\\
            TFMR + EN & \textbf{91.72} & \textbf{89.17} & \textbf{90.42} & 0.48 & 60.12 & 93.44 & 91.21\\
            TFMR + IN & 90.51 & 87.68 & 89.07 & 0.46 & 59.47 & 93.47 & 91.20\\
            TFMR + PRE + EN & 91.24 & 88.97 & 90.09 & 0.50 & 63.23 & 93.35 & 91.21\\
            TFMR + PRE + IN & 90.65 & 88.45 & 89.54 & 0.46 & 58.86 & \textbf{93.62} & \textbf{91.37} \\
            \hline
            \end{tabular} 
    }
    \vspace{-1em}
    \caption{Full UDS and English UD results for XLM-R models with and without pre-training.}
    \label{tab:bottom-up}
    \vspace{-1em}
\end{table}

The trends here suggest that a multilingual syntactic signal, when incorporated well into a UDS model, can provide benefit to the syntactic performance without necessarily reducing the semantic performance. 
Note that unlike in \S\ref{sec:experiment1}, the syntactic data used to pretrain the syntactic encoder and biaffine parser is neither parallel to the UDS dataset, nor is it in English. 
Thus, that the syntactic data can act as a signal for semantic parsing, albeit with small effects, is surprising. 

\vspace{-0.5em}
\paragraph{Top-down Effects} In the top-down direction (semantics to syntax) we train the encoder-side and intermediate variants of the joint UDS and syntactic parsing model and subsequently load the weights from their encoders and biaffine parsers into separate UD models for all 8 languages.
These are compared against a baseline model with weights initialized from a English UD parsing model.
Thus any improvement obtained by the encoder and intermediate models comes strictly from the semantic signal, since the syntactic signal is shared with the baseline model.
Table~\ref{tab:xlmr_syntax} shows the LAS and UAS performance of these models, with arrows indicating the direction of change from the baseline model. 
We see that almost all of the languages see benefits from the addition of semantic signal in at least one model variant, with the exception of Finnish, which performs worse across all variants and metrics when the semantic signal is included. 
For Galician, Hungarian, and Armenian, we see a sizeable improvement between the models. 
With the exception of Kazakh, whose train set is miniscule, these are among the lowest-resource languages considered. 
While, given the typical pipeline view of syntax as a substrate for semantics, we might expect the bottom-up results to be stronger than top-down results, here we find that the  syntactic benefits of pretraining on a semantic task are more consistent and stronger than in the other direction. 

\vspace{-0.5em} 
\paragraph{Discussion} On the whole, the multilingual top-down and bottom-up effects seem to mimick the monolingual results, albeit with smaller relative improvements. 
In \S\ref{sec:experiment1}, we saw a mutually beneficial relationship between UD and UDS parsing in a number of settings; here, we see that in several cases, this pattern generalizes to a case where we pretrain on data that is not only in a different domain than the evaluation (i.e. syntax vs. semantics) but also in a different language. 
These effects hint at useful commonalities not only between syntactic parses across multiple languages, but also between multilingual syntax and the UDS representation. 

\begin{table}[]
    \centering
\resizebox{\linewidth}{!}{%
    \begin{tabular}{ccccc}
        \hline
        Lang. & Base & Encoder & Intermediate \\
        \hline
AF & 86.10 / 89.06 & 86.40$\uparrow$ / 89.28$\uparrow$ & 85.27$\downarrow$ / 88.39$\downarrow$\\
DE & 84.30 / 88.45 & 84.23$\downarrow$ / 88.49$\uparrow$ & 83.49$\downarrow$ / 87.85$\downarrow$\\
FI & 80.94 / 85.83 & 80.53$\downarrow$ / 85.52$\downarrow$ & 80.50$\downarrow$ / 85.44$\downarrow$\\
FR & 86.66 / 90.02 & 86.82$\uparrow$ / 90.07$\uparrow$ & 86.71$\uparrow$ / 89.94$\downarrow$\\
GL & 78.60 / 82.46 & 79.58$\uparrow$ / 83.08$\uparrow$ & 79.77$\uparrow$ / 83.19$\uparrow$\\
HU & 82.90 / 87.76 & 83.67$\uparrow$ / 88.33$\uparrow$ & 82.01$\downarrow$ / 86.75$\downarrow$\\
HY & 77.97 / 83.72 & 79.31$\uparrow$ / 85.08$\uparrow$ & 76.48$\downarrow$ / 82.79$\downarrow$\\
KK & 51.05 / 67.09 & 51.25$\uparrow$ / 65.49$\downarrow$ & 48.85$\downarrow$ / 63.08$\downarrow$\\
\hline
AVG & 78.56 / 84.30 & 78.97$\uparrow$ / 84.42$\uparrow$ & 77.88$\downarrow$ / 83.43$\downarrow$\\
\hline
    \end{tabular}}
    \vspace{-1em}
    \caption{LAS/UAS for models with weights transferred from EWT UD parsing compared with those from encoder-side and intermediate UDS models.}
    \label{tab:xlmr_syntax}
    \vspace{-1.5em}
\end{table}

\section{Conclusion} In \S\ref{sec:syn_model}, we introduced a number of multitask architectures for joint syntactic and semantic parsing, which we demonstrated in \S\ref{sec:experiment1} can perform UD and UDS parsing simultaneously without sacrificing performance, as evaluated across a number of syntactic and semantic metrics. 
In particular, we observed a top-down benefit to syntactic parsing from the semantic signal as well as a bottom-up benefit to semantic performance from syntactic parsing. 
We contrasted both LSTM and Transformer-based variants of these architectures, finding the Transformer to be better on all metrics.
Finding the syntactic and semantic information present in the data to be complementary to that encoded in a frozen contextualized encoder, we experimented in \S\ref{sec:tuning} with tuning the encoder to varying depths, finding that tuning the top-most layers provides the greatest benefit. 
We analyzed the models resulting from this tuning step on their ability to resolve attachment ambiguities, as well as examining interactions between UDS annotations and UD dependency relations. 
Furthermore, in \S\ref{sec:multiling}, we extended our experiments beyond English, using a transfer-learning experimental paradigm to investigate effects between multilingual syntactic parsing in 8 languages and English semantic parsing, where we found similar trends to the English-only setting. 
Based on these multilingual results, we believe that expanding the UDS data paradigm (i.e. UD-based graph structure, continuous attributes) beyond English and building robust multilingual parsing models is a particularly promising direction for future work. 
Other directions include improving the robustness of the Transformer model in low-resource settings 
and improvements to the attribute modules. 

\section*{Acknowledgements} The first author is supported by an NSF Graduate Research Fellowship. This work was additionally supported by NSF grant 1763705, NSF grant 1749025, IARPA BETTER (2019-19051600005) and DARPA AIDA (FA8750-18-2-0015). We would like to thank the action editors, Carlos G\'{o}mez-Rodr\'{i}guez and Miguel Ballesteros, and the anonymous reviewers, whose thorough reading and thoughtful suggestions significantly improved our work. 

\bibliography{decomp_clean}

\begin{thebibliography}{63}
\expandafter\ifx\csname natexlab\endcsname\relax\def\natexlab#1{#1}\fi

\bibitem[{Abend et~al.(2017)Abend, Kwiatkowski, Smith, Goldwater, and
  Steedman}]{abend.o.2017}
Omri Abend, Tom Kwiatkowski, Nathaniel~J Smith, Sharon Goldwater, and Mark
  Steedman. 2017.
\newblock Bootstrapping language acquisition.
\newblock \emph{Cognition}, 164:116--143.

\bibitem[{Abend and Rappoport(2013)}]{abend.o.2013}
Omri Abend and Ari Rappoport. 2013.
\newblock \href {https://www.aclweb.org/anthology/P13-1023} {{U}niversal
  {C}onceptual {C}ognitive {A}nnotation ({UCCA})}.
\newblock In \emph{Proceedings of the 51st Annual Meeting of the Association
  for Computational Linguistics (Volume 1: Long Papers)}, pages 228--238,
  Sofia, Bulgaria. Association for Computational Linguistics.

\bibitem[{Artzi et~al.(2015)Artzi, Lee, and Zettlemoyer}]{artzi.y.2015}
Yoav Artzi, Kenton Lee, and Luke Zettlemoyer. 2015.
\newblock \href {https://doi.org/10.18653/v1/D15-1198} {Broad-coverage {CCG}
  semantic parsing with {AMR}}.
\newblock In \emph{Proceedings of the 2015 Conference on Empirical Methods in
  Natural Language Processing}, pages 1699--1710, Lisbon, Portugal. Association
  for Computational Linguistics.

\bibitem[{Banarescu et~al.(2013)Banarescu, Bonial, Cai, Georgescu, Griffitt,
  Hermjakob, Knight, Koehn, Palmer, and Schneider}]{banarescu.l.2013}
Laura Banarescu, Claire Bonial, Shu Cai, Madalina Georgescu, Kira Griffitt, Ulf
  Hermjakob, Kevin Knight, Philipp Koehn, Martha Palmer, and Nathan Schneider.
  2013.
\newblock \href {https://www.aclweb.org/anthology/W13-2322} {{A}bstract
  {M}eaning {R}epresentation for sembanking}.
\newblock In \emph{Proceedings of the 7th Linguistic Annotation Workshop and
  Interoperability with Discourse}, pages 178--186, Sofia, Bulgaria.
  Association for Computational Linguistics.

\bibitem[{Bergstra and Bengio(2012)}]{bergstra.j.2012}
James Bergstra and Yoshua Bengio. 2012.
\newblock Random search for hyper-parameter optimization.
\newblock \emph{The Journal of Machine Learning Research}, 13(1):281--305.

\bibitem[{Beschke(2019)}]{beschke.s.2019}
Sebastian Beschke. 2019.
\newblock \href {https://doi.org/10.26615/978-954-452-056-4_014} {Exploring
  graph-algebraic {CCG} combinators for syntactic-semantic {AMR} parsing}.
\newblock In \emph{Proceedings of the International Conference on Recent
  Advances in Natural Language Processing (RANLP 2019)}, pages 112--121, Varna,
  Bulgaria. INCOMA Ltd.

\bibitem[{Bies et~al.(2012)Bies, Mott, Warner, and Kulick}]{bies.a.2012}
Ann Bies, Justin Mott, Colin Warner, and Seth Kulick. 2012.
\newblock English {W}eb {T}reebank.
\newblock \emph{Linguistic Data Consortium, Philadelphia, PA}.

\bibitem[{Cai and Knight(2013)}]{cai.s.2013}
Shu Cai and Kevin Knight. 2013.
\newblock \href {https://www.aclweb.org/anthology/P13-2131} {{S}match: an
  evaluation metric for semantic feature structures}.
\newblock In \emph{Proceedings of the 51st Annual Meeting of the Association
  for Computational Linguistics (Volume 2: Short Papers)}, pages 748--752,
  Sofia, Bulgaria. Association for Computational Linguistics.

\bibitem[{Caruana(1997)}]{caruana.r.1997}
Rich Caruana. 1997.
\newblock Multitask learning.
\newblock \emph{Machine learning}, 28(1):41--75.

\bibitem[{Chen and Manning(2014)}]{chen.d.2014}
Danqi Chen and Christopher Manning. 2014.
\newblock \href {https://doi.org/10.3115/v1/D14-1082} {A fast and accurate
  dependency parser using neural networks}.
\newblock In \emph{Proceedings of the 2014 Conference on Empirical Methods in
  Natural Language Processing ({EMNLP})}, pages 740--750, Doha, Qatar.
  Association for Computational Linguistics.

\bibitem[{Chomsky(1969)}]{chomsky.c.1969}
Carol Chomsky. 1969.
\newblock The acquisition of syntax in children from 5 to 10.

\bibitem[{Chu(1965)}]{chu.l.1965}
Yoeng-Jin Chu. 1965.
\newblock On the shortest arborescence of a directed graph.
\newblock \emph{Scientia Sinica}, 14:1396--1400.

\bibitem[{Conneau et~al.(2020)Conneau, Khandelwal, Goyal, Chaudhary, Wenzek,
  Guzm{\'a}n, Grave, Ott, Zettlemoyer, and Stoyanov}]{conneau.a.2019}
Alexis Conneau, Kartikay Khandelwal, Naman Goyal, Vishrav Chaudhary, Guillaume
  Wenzek, Francisco Guzm{\'a}n, Edouard Grave, Myle Ott, Luke Zettlemoyer, and
  Veselin Stoyanov. 2020.
\newblock \href {https://doi.org/10.18653/v1/2020.acl-main.747} {Unsupervised
  cross-lingual representation learning at scale}.
\newblock In \emph{Proceedings of the 58th Annual Meeting of the Association
  for Computational Linguistics}, pages 8440--8451, Online. Association for
  Computational Linguistics.

\bibitem[{Cromer(1970)}]{cromer.r.1970}
Richard~F Cromer. 1970.
\newblock ‘{C}hildren are nice to understand’: Surface structure clues for
  the recovery of a deep structure.
\newblock \emph{British Journal of Psychology}, 61(3):397--408.

\bibitem[{Devlin(2018)}]{devlin.j.2018}
Jacob Devlin. 2018.
\newblock \href
  {https://github.com/google-research/bert/blob/a9ba4b8d7704c1ae18d1b28c56c0430d41407eb1/multilingual.md}
  {\emph{Multilingual {BERT} {README} document.}}

\bibitem[{Devlin et~al.(2019)Devlin, Chang, Lee, and Toutanova}]{devlin.j.2019}
Jacob Devlin, Ming-Wei Chang, Kenton Lee, and Kristina Toutanova. 2019.
\newblock \href {https://doi.org/10.18653/v1/N19-1423} {{BERT}: Pre-training of
  deep bidirectional transformers for language understanding}.
\newblock In \emph{Proceedings of the 2019 Conference of the North {A}merican
  Chapter of the Association for Computational Linguistics: Human Language
  Technologies, Volume 1 (Long and Short Papers)}, pages 4171--4186,
  Minneapolis, Minnesota. Association for Computational Linguistics.

\bibitem[{Dowty(1991)}]{dowty.d.1991}
David Dowty. 1991.
\newblock Thematic proto-roles and argument selection.
\newblock \emph{Language}, 67(3):547--619.

\bibitem[{Dozat and Manning(2017)}]{dozat.t.2016}
Timothy Dozat and Christopher~D. Manning. 2017.
\newblock \href {https://openreview.net/forum?id=Hk95PK9le} {Deep biaffine
  attention for neural dependency parsing}.
\newblock In \emph{5th International Conference on Learning Representations,
  {ICLR} 2017, Toulon, France, April 24-26, 2017, Conference Track
  Proceedings}. OpenReview.net.

\bibitem[{Edmonds(1967)}]{edmonds.j.1967}
Jack Edmonds. 1967.
\newblock Optimum branchings.
\newblock \emph{Journal of Research of the national Bureau of Standards B},
  71(4):233--240.

\bibitem[{Gardner et~al.(2020)Gardner, Artzi, Basmov, Berant, Bogin, Chen,
  Dasigi, Dua, Elazar, Gottumukkala, Gupta, Hajishirzi, Ilharco, Khashabi, Lin,
  Liu, Liu, Mulcaire, Ning, Singh, Smith, Subramanian, Tsarfaty, Wallace,
  Zhang, and Zhou}]{gardner.m.2020}
Matt Gardner, Yoav Artzi, Victoria Basmov, Jonathan Berant, Ben Bogin, Sihao
  Chen, Pradeep Dasigi, Dheeru Dua, Yanai Elazar, Ananth Gottumukkala, Nitish
  Gupta, Hannaneh Hajishirzi, Gabriel Ilharco, Daniel Khashabi, Kevin Lin,
  Jiangming Liu, Nelson~F. Liu, Phoebe Mulcaire, Qiang Ning, Sameer Singh,
  Noah~A. Smith, Sanjay Subramanian, Reut Tsarfaty, Eric Wallace, Ally Zhang,
  and Ben Zhou. 2020.
\newblock \href {https://doi.org/10.18653/v1/2020.findings-emnlp.117}
  {Evaluating models{'} local decision boundaries via contrast sets}.
\newblock In \emph{Findings of the Association for Computational Linguistics:
  EMNLP 2020}, pages 1307--1323, Online. Association for Computational
  Linguistics.

\bibitem[{Gardner et~al.(2018)Gardner, Grus, Neumann, Tafjord, Dasigi, Liu,
  Peters, Schmitz, and Zettlemoyer}]{gardner.m.2017}
Matt Gardner, Joel Grus, Mark Neumann, Oyvind Tafjord, Pradeep Dasigi,
  Nelson~F. Liu, Matthew Peters, Michael Schmitz, and Luke Zettlemoyer. 2018.
\newblock \href {https://doi.org/10.18653/v1/W18-2501} {{A}llen{NLP}: A deep
  semantic natural language processing platform}.
\newblock In \emph{Proceedings of Workshop for {NLP} Open Source Software
  ({NLP}-{OSS})}, pages 1--6, Melbourne, Australia. Association for
  Computational Linguistics.

\bibitem[{Glava{\v{s}} and Vuli{\'c}(2020)}]{glavas.g.2020}
Goran Glava{\v{s}} and Ivan Vuli{\'c}. 2020.
\newblock Is supervised syntactic parsing beneficial for language
  understanding? an empirical investigation.
\newblock \emph{arXiv preprint arXiv:2008.06788}.

\bibitem[{Gleitman(1990)}]{gleitman.l.1990}
Lila Gleitman. 1990.
\newblock The structural sources of verb meanings.
\newblock \emph{Language acquisition}, 1(1):3--55.

\bibitem[{Haji{\v{c}} et~al.(2009)Haji{\v{c}}, Ciaramita, Johansson, Kawahara,
  Mart{\'\i}, M{\`a}rquez, Meyers, Nivre, Pad{\'o}, {\v{S}}t{\v{e}}p{\'a}nek,
  Stra{\v{n}}{\'a}k, Surdeanu, Xue, and Zhang}]{hajivc.j.2009}
Jan Haji{\v{c}}, Massimiliano Ciaramita, Richard Johansson, Daisuke Kawahara,
  Maria~Ant{\`o}nia Mart{\'\i}, Llu{\'\i}s M{\`a}rquez, Adam Meyers, Joakim
  Nivre, Sebastian Pad{\'o}, Jan {\v{S}}t{\v{e}}p{\'a}nek, Pavel
  Stra{\v{n}}{\'a}k, Mihai Surdeanu, Nianwen Xue, and Yi~Zhang. 2009.
\newblock \href {https://www.aclweb.org/anthology/W09-1201} {The {C}o{NLL}-2009
  shared task: Syntactic and semantic dependencies in multiple languages}.
\newblock In \emph{Proceedings of the Thirteenth Conference on Computational
  Natural Language Learning ({C}o{NLL} 2009): Shared Task}, pages 1--18,
  Boulder, Colorado. Association for Computational Linguistics.

\bibitem[{Hewitt and Liang(2019)}]{hewitt.j.2019}
John Hewitt and Percy Liang. 2019.
\newblock \href {https://doi.org/10.18653/v1/D19-1275} {Designing and
  interpreting probes with control tasks}.
\newblock In \emph{Proceedings of the 2019 Conference on Empirical Methods in
  Natural Language Processing and the 9th International Joint Conference on
  Natural Language Processing (EMNLP-IJCNLP)}, pages 2733--2743, Hong Kong,
  China. Association for Computational Linguistics.

\bibitem[{Jawahar et~al.(2019)Jawahar, Sagot, and Seddah}]{jawahar.g.2019}
Ganesh Jawahar, Beno{\^\i}t Sagot, and Djam{\'e} Seddah. 2019.
\newblock \href {https://doi.org/10.18653/v1/P19-1356} {What does {BERT} learn
  about the structure of language?}
\newblock In \emph{Proceedings of the 57th Annual Meeting of the Association
  for Computational Linguistics}, pages 3651--3657, Florence, Italy.
  Association for Computational Linguistics.

\bibitem[{Johansson and Nugues(2008)}]{johansson.r.2008}
Richard Johansson and Pierre Nugues. 2008.
\newblock \href {https://www.aclweb.org/anthology/W08-2123} {Dependency-based
  syntactic{--}semantic analysis with {P}rop{B}ank and {N}om{B}ank}.
\newblock In \emph{{C}o{NLL} 2008: Proceedings of the Twelfth Conference on
  Computational Natural Language Learning}, pages 183--187, Manchester,
  England. Coling 2008 Organizing Committee.

\bibitem[{Kondratyuk and Straka(2019)}]{kondratyuk.d.2019}
Dan Kondratyuk and Milan Straka. 2019.
\newblock \href {https://doi.org/10.18653/v1/D19-1279} {75 languages, 1 model:
  Parsing {U}niversal {D}ependencies universally}.
\newblock In \emph{Proceedings of the 2019 Conference on Empirical Methods in
  Natural Language Processing and the 9th International Joint Conference on
  Natural Language Processing (EMNLP-IJCNLP)}, pages 2779--2795, Hong Kong,
  China. Association for Computational Linguistics.

\bibitem[{Krishnamurthy and Mitchell(2014)}]{krishnamurthy.j.2014}
Jayant Krishnamurthy and Tom~M. Mitchell. 2014.
\newblock \href {https://doi.org/10.3115/v1/P14-1112} {Joint syntactic and
  semantic parsing with {C}ombinatory {C}ategorial {G}rammar}.
\newblock In \emph{Proceedings of the 52nd Annual Meeting of the Association
  for Computational Linguistics (Volume 1: Long Papers)}, pages 1188--1198,
  Baltimore, Maryland. Association for Computational Linguistics.

\bibitem[{Landau and Gleitman(1985)}]{landau.b.1985}
Barbara Landau and Lila~R. Gleitman. 1985.
\newblock \emph{Language and Experience: {Evidence} from the Blind Child}.
\newblock Number~8 in Cognitive Science Series. Harvard University Press,
  Cambridge, MA.

\bibitem[{Lewis et~al.(2015)Lewis, He, and Zettlemoyer}]{lewis.m.2015}
Mike Lewis, Luheng He, and Luke Zettlemoyer. 2015.
\newblock \href {https://doi.org/10.18653/v1/D15-1169} {Joint {A}* {CCG}
  parsing and semantic role labelling}.
\newblock In \emph{Proceedings of the 2015 Conference on Empirical Methods in
  Natural Language Processing}, pages 1444--1454, Lisbon, Portugal. Association
  for Computational Linguistics.

\bibitem[{Misra and Artzi(2016)}]{misra.d.2016}
Dipendra~Kumar Misra and Yoav Artzi. 2016.
\newblock \href {https://doi.org/10.18653/v1/D16-1183} {Neural shift-reduce
  {CCG} semantic parsing}.
\newblock In \emph{Proceedings of the 2016 Conference on Empirical Methods in
  Natural Language Processing}, pages 1775--1786, Austin, Texas. Association
  for Computational Linguistics.

\bibitem[{Montague(1970)}]{montague.r.1970}
Richard Montague. 1970.
\newblock Universal grammar.
\newblock \emph{Theoria}, 36(3):373--398.

\bibitem[{Naigles(1990)}]{naigles.l.1990}
Letitia Naigles. 1990.
\newblock Children use syntax to learn verb meanings.
\newblock \emph{Journal of child language}, 17(2):357--374.

\bibitem[{Nguyen and Salazar(2019)}]{nguyen.t.2019}
Toan~Q Nguyen and Julian Salazar. 2019.
\newblock Transformers without tears: Improving the normalization of
  self-attention.
\newblock \emph{arXiv preprint arXiv:1910.05895}.

\bibitem[{Oepen et~al.(2016)Oepen, Kuhlmann, Miyao, Zeman, Cinkov{\'a},
  Flickinger, Haji{\v{c}}, Ivanova, and Ure{\v{s}}ov{\'a}}]{oepen.s.2016}
Stephan Oepen, Marco Kuhlmann, Yusuke Miyao, Daniel Zeman, Silvie Cinkov{\'a},
  Dan Flickinger, Jan Haji{\v{c}}, Angelina Ivanova, and Zde{\v{n}}ka
  Ure{\v{s}}ov{\'a}. 2016.
\newblock \href {https://www.aclweb.org/anthology/L16-1630} {Towards
  comparability of linguistic graph {B}anks for semantic parsing}.
\newblock In \emph{Proceedings of the Tenth International Conference on
  Language Resources and Evaluation ({LREC}'16)}, pages 3991--3995,
  Portoro{\v{z}}, Slovenia. European Language Resources Association (ELRA).

\bibitem[{Oepen et~al.(2014{\natexlab{a}})Oepen, Kuhlmann, Miyao, Zeman,
  Flickinger, Haji{\v{c}}, Ivanova, and Zhang}]{oepen.s.2014}
Stephan Oepen, Marco Kuhlmann, Yusuke Miyao, Daniel Zeman, Dan Flickinger, Jan
  Haji{\v{c}}, Angelina Ivanova, and Yi~Zhang. 2014{\natexlab{a}}.
\newblock \href {https://doi.org/10.3115/v1/S14-2008} {{S}em{E}val 2014 task 8:
  Broad-coverage semantic dependency parsing}.
\newblock In \emph{Proceedings of the 8th International Workshop on Semantic
  Evaluation ({S}em{E}val 2014)}, pages 63--72, Dublin, Ireland. Association
  for Computational Linguistics.

\bibitem[{Oepen et~al.(2014{\natexlab{b}})Oepen, Kuhlmann, Miyao, Zeman,
  Flickinger, Haji{\v{c}}, Ivanova, and Zhang}]{oepen.s.2015}
Stephan Oepen, Marco Kuhlmann, Yusuke Miyao, Daniel Zeman, Dan Flickinger, Jan
  Haji{\v{c}}, Angelina Ivanova, and Yi~Zhang. 2014{\natexlab{b}}.
\newblock \href {https://doi.org/10.3115/v1/S14-2008} {{S}em{E}val 2014 task 8:
  Broad-coverage semantic dependency parsing}.
\newblock In \emph{Proceedings of the 8th International Workshop on Semantic
  Evaluation ({S}em{E}val 2014)}, pages 63--72, Dublin, Ireland. Association
  for Computational Linguistics.

\bibitem[{Pinker(1984)}]{pinker.s.1984}
S~Pinker. 1984.
\newblock 1984: Language learnability and language development. {C}ambridge,
  {MA}: Harvard {U}niversity {P}ress.

\bibitem[{Pinker(1979)}]{pinker.s.1979}
Steven Pinker. 1979.
\newblock \href {https://doi.org/https://doi.org/10.1016/0010-0277(79)90001-5}
  {Formal models of language learning}.
\newblock \emph{Cognition}, 7(3):217 -- 283.

\bibitem[{Pollard and Sag(1994)}]{pollard.c.1994}
Carl Pollard and Ivan~A Sag. 1994.
\newblock \emph{Head-driven phrase structure grammar}.
\newblock University of Chicago Press.

\bibitem[{Schuster et~al.(2019)Schuster, Ram, Barzilay, and
  Globerson}]{schuster.t.2019}
Tal Schuster, Ori Ram, Regina Barzilay, and Amir Globerson. 2019.
\newblock \href {https://doi.org/10.18653/v1/N19-1162} {Cross-lingual alignment
  of contextual word embeddings, with applications to zero-shot dependency
  parsing}.
\newblock In \emph{Proceedings of the 2019 Conference of the North {A}merican
  Chapter of the Association for Computational Linguistics: Human Language
  Technologies, Volume 1 (Long and Short Papers)}, pages 1599--1613,
  Minneapolis, Minnesota. Association for Computational Linguistics.

\bibitem[{See et~al.(2017)See, Liu, and Manning}]{see.a.2017}
Abigail See, Peter~J. Liu, and Christopher~D. Manning. 2017.
\newblock \href {https://doi.org/10.18653/v1/P17-1099} {Get to the point:
  Summarization with pointer-generator networks}.
\newblock In \emph{Proceedings of the 55th Annual Meeting of the Association
  for Computational Linguistics (Volume 1: Long Papers)}, pages 1073--1083,
  Vancouver, Canada. Association for Computational Linguistics.

\bibitem[{Steedman(2000)}]{steedman.m.2000}
Mark Steedman. 2000.
\newblock \emph{The syntactic process}, volume~24.
\newblock MIT press Cambridge, MA.

\bibitem[{Stengel-Eskin et~al.(2020)Stengel-Eskin, White, Zhang, and
  Van~Durme}]{stengel-eskin.e.2020}
Elias Stengel-Eskin, Aaron~Steven White, Sheng Zhang, and Benjamin Van~Durme.
  2020.
\newblock \href {https://doi.org/10.18653/v1/2020.acl-main.746} {Universal
  decompositional semantic parsing}.
\newblock In \emph{Proceedings of the 58th Annual Meeting of the Association
  for Computational Linguistics}, pages 8427--8439, Online. Association for
  Computational Linguistics.

\bibitem[{Strubell et~al.(2018)Strubell, Verga, Andor, Weiss, and
  McCallum}]{strubell.e.2018}
Emma Strubell, Patrick Verga, Daniel Andor, David Weiss, and Andrew McCallum.
  2018.
\newblock \href {https://doi.org/10.18653/v1/D18-1548} {Linguistically-informed
  self-attention for semantic role labeling}.
\newblock In \emph{Proceedings of the 2018 Conference on Empirical Methods in
  Natural Language Processing}, pages 5027--5038, Brussels, Belgium.
  Association for Computational Linguistics.

\bibitem[{Surdeanu et~al.(2008)Surdeanu, Johansson, Meyers, M{\`a}rquez, and
  Nivre}]{surdeanu.m.2008}
Mihai Surdeanu, Richard Johansson, Adam Meyers, Llu{\'\i}s M{\`a}rquez, and
  Joakim Nivre. 2008.
\newblock \href {https://www.aclweb.org/anthology/W08-2121} {The {C}o{NLL} 2008
  shared task on joint parsing of syntactic and semantic dependencies}.
\newblock In \emph{{C}o{NLL} 2008: Proceedings of the Twelfth Conference on
  Computational Natural Language Learning}, pages 159--177, Manchester,
  England. Coling 2008 Organizing Committee.

\bibitem[{Swayamdipta et~al.(2016)Swayamdipta, Ballesteros, Dyer, and
  Smith}]{swayamdipta.s.2016}
Swabha Swayamdipta, Miguel Ballesteros, Chris Dyer, and Noah~A. Smith. 2016.
\newblock \href {https://doi.org/10.18653/v1/K16-1019} {Greedy, joint
  syntactic-semantic parsing with stack {LSTM}s}.
\newblock In \emph{Proceedings of The 20th {SIGNLL} Conference on Computational
  Natural Language Learning}, pages 187--197, Berlin, Germany. Association for
  Computational Linguistics.

\bibitem[{Swayamdipta et~al.(2019)Swayamdipta, Peters, Roof, Dyer, and
  Smith}]{swayamdipta.s.2019}
Swabha Swayamdipta, Matthew Peters, Brendan Roof, Chris Dyer, and Noah~A Smith.
  2019.
\newblock Shallow syntax in deep water.
\newblock \emph{arXiv preprint arXiv:1908.11047}.

\bibitem[{Swayamdipta et~al.(2017)Swayamdipta, Thomson, Dyer, and
  Smith}]{swayamdipta.s.2017}
Swabha Swayamdipta, Sam Thomson, Chris Dyer, and Noah~A Smith. 2017.
\newblock Frame-semantic parsing with softmax-margin segmental rnns and a
  syntactic scaffold.
\newblock \emph{arXiv preprint arXiv:1706.09528}.

\bibitem[{Swayamdipta et~al.(2018)Swayamdipta, Thomson, Lee, Zettlemoyer, Dyer,
  and Smith}]{swayamdipta.s.2018}
Swabha Swayamdipta, Sam Thomson, Kenton Lee, Luke Zettlemoyer, Chris Dyer, and
  Noah~A. Smith. 2018.
\newblock \href {https://doi.org/10.18653/v1/D18-1412} {Syntactic scaffolds for
  semantic structures}.
\newblock In \emph{Proceedings of the 2018 Conference on Empirical Methods in
  Natural Language Processing}, pages 3772--3782, Brussels, Belgium.
  Association for Computational Linguistics.

\bibitem[{Tenney et~al.(2019)Tenney, Das, and Pavlick}]{tenney.i.2019}
Ian Tenney, Dipanjan Das, and Ellie Pavlick. 2019.
\newblock \href {https://doi.org/10.18653/v1/P19-1452} {{BERT} rediscovers the
  classical {NLP} pipeline}.
\newblock In \emph{Proceedings of the 57th Annual Meeting of the Association
  for Computational Linguistics}, pages 4593--4601, Florence, Italy.
  Association for Computational Linguistics.

\bibitem[{Vaswani et~al.(2017)Vaswani, Shazeer, Parmar, Uszkoreit, Jones,
  Gomez, Kaiser, and Polosukhin}]{vaswani.a.2017}
Ashish Vaswani, Noam Shazeer, Niki Parmar, Jakob Uszkoreit, Llion Jones,
  Aidan~N Gomez, {\L}ukasz Kaiser, and Illia Polosukhin. 2017.
\newblock Attention is all you need.
\newblock In \emph{Advances in neural information processing systems}, pages
  5998--6008.

\bibitem[{White(2020)}]{white.a.2020b}
Aaron~Steven White. 2020.
\newblock Lexically triggered veridicality inferences.
\newblock \emph{Handbook of Pragmatics: 22nd Annual Installment}, 22:115.

\bibitem[{White et~al.(2016)White, Reisinger, Sakaguchi, Vieira, Zhang,
  Rudinger, Rawlins, and Van~Durme}]{white.a.2016}
Aaron~Steven White, Drew Reisinger, Keisuke Sakaguchi, Tim Vieira, Sheng Zhang,
  Rachel Rudinger, Kyle Rawlins, and Benjamin Van~Durme. 2016.
\newblock \href {https://doi.org/10.18653/v1/D16-1177} {Universal
  decompositional semantics on {U}niversal {D}ependencies}.
\newblock In \emph{Proceedings of the 2016 Conference on Empirical Methods in
  Natural Language Processing}, pages 1713--1723, Austin, Texas. Association
  for Computational Linguistics.

\bibitem[{White et~al.(2018)White, Rudinger, Rawlins, and
  Van~Durme}]{white.a.2018}
Aaron~Steven White, Rachel Rudinger, Kyle Rawlins, and Benjamin Van~Durme.
  2018.
\newblock \href {https://doi.org/10.18653/v1/D18-1501} {Lexicosyntactic
  inference in neural models}.
\newblock In \emph{Proceedings of the 2018 Conference on Empirical Methods in
  Natural Language Processing}, pages 4717--4724, Brussels, Belgium.
  Association for Computational Linguistics.

\bibitem[{White et~al.(2020)White, Stengel-Eskin, Vashishtha, Govindarajan,
  Reisinger, Vieira, Sakaguchi, Zhang, Ferraro, Rudinger, Rawlins, and
  Van~Durme}]{white.a.2020}
Aaron~Steven White, Elias Stengel-Eskin, Siddharth Vashishtha,
  Venkata~Subrahmanyan Govindarajan, Dee~Ann Reisinger, Tim Vieira, Keisuke
  Sakaguchi, Sheng Zhang, Francis Ferraro, Rachel Rudinger, Kyle Rawlins, and
  Benjamin Van~Durme. 2020.
\newblock \href {https://www.aclweb.org/anthology/2020.lrec-1.699} {The
  universal decompositional semantics dataset and decomp toolkit}.
\newblock In \emph{Proceedings of the 12th Language Resources and Evaluation
  Conference}, pages 5698--5707, Marseille, France. European Language Resources
  Association.

\bibitem[{Zeman et~al.(2018)Zeman, Haji{\v{c}}, Popel, Potthast, Straka,
  Ginter, Nivre, and Petrov}]{zeman.d.2018}
Daniel Zeman, Jan Haji{\v{c}}, Martin Popel, Martin Potthast, Milan Straka,
  Filip Ginter, Joakim Nivre, and Slav Petrov. 2018.
\newblock \href {https://doi.org/10.18653/v1/K18-2001} {{C}o{NLL} 2018 shared
  task: Multilingual parsing from raw text to {U}niversal {D}ependencies}.
\newblock In \emph{Proceedings of the {C}o{NLL} 2018 Shared Task: Multilingual
  Parsing from Raw Text to Universal Dependencies}, pages 1--21, Brussels,
  Belgium. Association for Computational Linguistics.

\bibitem[{Zeman et~al.(2017)Zeman, Popel, Straka, Haji{\v{c}}, Nivre, Ginter,
  Luotolahti, Pyysalo, Petrov, Potthast, Tyers, Badmaeva, Gokirmak, Nedoluzhko,
  Cinkov{\'a}, Haji{\v{c}}~jr., Hlav{\'a}{\v{c}}ov{\'a}, Kettnerov{\'a},
  Ure{\v{s}}ov{\'a}, Kanerva, Ojala, Missil{\"a}, Manning, Schuster, Reddy,
  Taji, Habash, Leung, de~Marneffe, Sanguinetti, Simi, Kanayama, de~Paiva,
  Droganova, Mart{\'\i}nez~Alonso, {\c{C}}{\"o}ltekin, Sulubacak, Uszkoreit,
  Macketanz, Burchardt, Harris, Marheinecke, Rehm, Kayadelen, Attia, Elkahky,
  Yu, Pitler, Lertpradit, Mandl, Kirchner, Alcalde, Strnadov{\'a}, Banerjee,
  Manurung, Stella, Shimada, Kwak, Mendon{\c{c}}a, Lando, Nitisaroj, and
  Li}]{zeman.d.2017}
Daniel Zeman, Martin Popel, Milan Straka, Jan Haji{\v{c}}, Joakim Nivre, Filip
  Ginter, Juhani Luotolahti, Sampo Pyysalo, Slav Petrov, Martin Potthast,
  Francis Tyers, Elena Badmaeva, Memduh Gokirmak, Anna Nedoluzhko, Silvie
  Cinkov{\'a}, Jan Haji{\v{c}}~jr., Jaroslava Hlav{\'a}{\v{c}}ov{\'a},
  V{\'a}clava Kettnerov{\'a}, Zde{\v{n}}ka Ure{\v{s}}ov{\'a}, Jenna Kanerva,
  Stina Ojala, Anna Missil{\"a}, Christopher~D. Manning, Sebastian Schuster,
  Siva Reddy, Dima Taji, Nizar Habash, Herman Leung, Marie-Catherine
  de~Marneffe, Manuela Sanguinetti, Maria Simi, Hiroshi Kanayama, Valeria
  de~Paiva, Kira Droganova, H{\'e}ctor Mart{\'\i}nez~Alonso,
  {\c{C}}a{\u{g}}r{\i} {\c{C}}{\"o}ltekin, Umut Sulubacak, Hans Uszkoreit,
  Vivien Macketanz, Aljoscha Burchardt, Kim Harris, Katrin Marheinecke, Georg
  Rehm, Tolga Kayadelen, Mohammed Attia, Ali Elkahky, Zhuoran Yu, Emily Pitler,
  Saran Lertpradit, Michael Mandl, Jesse Kirchner, Hector~Fernandez Alcalde,
  Jana Strnadov{\'a}, Esha Banerjee, Ruli Manurung, Antonio Stella, Atsuko
  Shimada, Sookyoung Kwak, Gustavo Mendon{\c{c}}a, Tatiana Lando, Rattima
  Nitisaroj, and Josie Li. 2017.
\newblock \href {https://doi.org/10.18653/v1/K17-3001} {{C}o{NLL} 2017 shared
  task: Multilingual parsing from raw text to {U}niversal {D}ependencies}.
\newblock In \emph{Proceedings of the {C}o{NLL} 2017 Shared Task: Multilingual
  Parsing from Raw Text to Universal Dependencies}, pages 1--19, Vancouver,
  Canada. Association for Computational Linguistics.

\bibitem[{Zhang et~al.(2019{\natexlab{a}})Zhang, Ma, Duh, and
  Van~Durme}]{zhang.s.2019a}
Sheng Zhang, Xutai Ma, Kevin Duh, and Benjamin Van~Durme. 2019{\natexlab{a}}.
\newblock \href {https://doi.org/10.18653/v1/P19-1009} {{AMR} parsing as
  sequence-to-graph transduction}.
\newblock In \emph{Proceedings of the 57th Annual Meeting of the Association
  for Computational Linguistics}, pages 80--94, Florence, Italy. Association
  for Computational Linguistics.

\bibitem[{Zhang et~al.(2019{\natexlab{b}})Zhang, Ma, Duh, and
  Van~Durme}]{zhang.s.2019b}
Sheng Zhang, Xutai Ma, Kevin Duh, and Benjamin Van~Durme. 2019{\natexlab{b}}.
\newblock \href {https://doi.org/10.18653/v1/D19-1392} {Broad-coverage semantic
  parsing as transduction}.
\newblock In \emph{Proceedings of the 2019 Conference on Empirical Methods in
  Natural Language Processing and the 9th International Joint Conference on
  Natural Language Processing (EMNLP-IJCNLP)}, pages 3786--3798, Hong Kong,
  China. Association for Computational Linguistics.

\bibitem[{Zhang et~al.(2017)Zhang, Rudinger, and Van~Durme}]{zhang.s.2017}
Sheng Zhang, Rachel Rudinger, and Benjamin Van~Durme. 2017.
\newblock \href {https://www.aclweb.org/anthology/W17-6944} {An evaluation of
  {P}red{P}att and open {IE} via stage 1 semantic role labeling}.
\newblock In \emph{{IWCS} 2017 {---} 12th International Conference on
  Computational Semantics {---} Short papers}.

\bibitem[{Zhou et~al.(2020)Zhou, Zhang, Ji, and Tang}]{zhou.q.2020}
Qiji Zhou, Yue Zhang, Donghong Ji, and Hao Tang. 2020.
\newblock \href {https://doi.org/10.18653/v1/2020.acl-main.397} {{AMR} parsing
  with latent structural information}.
\newblock In \emph{Proceedings of the 58th Annual Meeting of the Association
  for Computational Linguistics}, pages 4306--4319, Online. Association for
  Computational Linguistics.

\end{thebibliography}
\bibliographystyle{acl_natbib}

\end{document}